\documentclass[letterpaper,11pt]{article}
\usepackage[in]{fullpage}
\usepackage{amsmath,amsthm,amsfonts,mathtools}
\usepackage{booktabs}
\usepackage{amssymb,mathrsfs,array,tabularx,flafter}
\usepackage{xurl}
\usepackage{tikz}
\usepackage{enumitem}
\usepackage[ruled,vlined,linesnumbered]{algorithm2e}
\PrintSemicolon
\SetArgSty{textup}
\usepackage{hyperref}
\hypersetup{hidelinks,pdftitle={Beyond Monotonicity: Stochastic Online Optimization under Semi-Bandit Feedback},pdfauthor={Tsubasa Harada, Yasushi Kawase, Hanna Sumita}}
\allowdisplaybreaks[2]
\numberwithin{equation}{section}
\newtheorem{theorem}{Theorem}[section]
\newtheorem{lemma}[theorem]{Lemma}
\newtheorem{assumption}[theorem]{Assumption}
\newtheorem{proposition}[theorem]{Proposition}
\newtheorem{corollary}[theorem]{Corollary}

\newcommand{\E}{\mathbb E}
\newcommand{\Prb}{\mathbb P}
\newcommand{\Var}{\operatorname{Var}}
\newcommand{\TV}{d_{\mathrm{TV}}}
\newcommand{\opt}{\mathrm{OPT}}
\newcommand{\KL}{\mathrm{KL}}
\newcommand{\obs}{\mathrm{obs}}
\newcommand{\child}{\mathrm{ch}}

\newcommand{\reg}{\mathrm{Reg}}
\newcommand{\ind}{\mathbf 1}
\newcommand{\PP}{\mathcal P}
\newcommand{\CC}{\mathcal C}
\newcommand{\II}{\mathcal I}
\newcommand{\FF}{\mathcal F}

\newcommand{\otil}{\widetilde O}
\newcommand{\thetatil}{\widetilde\Theta}
\title{Beyond Monotonicity: Stochastic Online Optimization \\ under Semi-Bandit Feedback}
\author{Tsubasa Harada\\Institute of Science Tokyo
\and Yasushi Kawase\\Chuo University
\and Hanna Sumita\\Institute of Science Tokyo}
\date{}
\begin{document}
\pagenumbering{roman}
\maketitle
\thispagestyle{empty}
\vspace{-7mm}
\begin{abstract}
We study stochastic online optimization (SOO) under semi-bandit feedback, in which a policy adaptively observes $n$ independent random variables with unknown distributions in each of $T$ rounds, and only the observed values are available for learning.
Monotonicity with respect to first-order stochastic dominance enables general reductions for learning SOO, but it fails in natural problems where observations change future feasibility, resource use, or reference points.
We provide an algorithm that runs the dynamic program for the optimal policy with each expectation replaced by its most optimistic value over a confidence set of distributions.
Its regret is governed by the level-set complexity $s_i$ of the values computed by this dynamic program as functions of the $i$th variable, that is, the number of intervals needed to describe their superlevel sets.
For rewards in $[0,1]$, the regret is $\otil(\sqrt{ST})$ without this monotonicity, where $S=\sum_{i=1}^n\min\{s_i,k_i\}$ and $k_i$ is the support size of the $i$th distribution, which may be infinite when $s_i$ is finite.
In particular, when the level-set complexities are bounded by a constant, the regret is $\otil(\sqrt{nT})$, even for continuous distributions.
The algorithm requires no knowledge of $s_i$, $k_i$, or the supports, and it also achieves regret $\otil(\sqrt{nkT})$ for supports of size at most $k$ without any level-set condition, which improves the previous bound $\otil(n(\sqrt{kT}+k))$ for FSD-monotone problems.
We also prove a lower bound of $\Omega(\min\{T,\sqrt{ST}\})$ for all positive integers $s_1,\ldots,s_n$ with $k_i=2s_i+1$, which holds even under full feedback and matches the upper bound up to logarithmic factors.
As applications, we obtain regret $\otil(\sqrt{nT})$ for the following problems, none of which is monotone under stochastic dominance: stochastic Boolean function evaluation, the full-information secretary problem, optimal stopping with loss aversion, scheduling with deadlines and correlated stochastic knapsack with unimodal rewards, and overbooking with group reservations.
\end{abstract}
\clearpage
\pagenumbering{arabic}

\section{Introduction}\label{sec:intro}
Stochastic online optimization (SOO) is a framework for sequential decision-making under uncertainty.
The input consists of $n$ independent random variables, and a \emph{policy} adaptively chooses the next action based on the values observed so far.
This framework includes optimal stopping problems such as Pandora's box~\cite{weitzman1979} and prophet inequalities~\cite{samuel1984comparison,kleinberg2012matroid}, as well as stochastic probing~\cite{GN13}, stochastic submodular optimization~\cite{asadpour2016maximizing}, and other stochastic combinatorial optimization problems.

Ghuge, Zhuo, Agarwal, and Nagarajan~\cite{ghuge2026semibandit} (a preliminary version appeared at FOCS 2024) introduced a general framework for learning SOO when the distributions are unknown, in which the same instance is played for $T$ rounds.
In each round, the learner chooses a policy and receives \emph{semi-bandit feedback}, which consists of the values of the variables observed during the execution of the policy.
The goal is to minimize the \emph{regret}, which is the cumulative loss in expected reward compared with the optimal policy that knows the distributions.
Their framework assumes monotonicity with respect to first-order stochastic dominance (FSD), called \emph{FSD monotonicity}, which means that replacing each distribution by a stochastically dominating one never decreases (or never increases) the optimal expected reward.
Under this assumption, running an algorithm for known distributions on optimistic distributions that dominate the true ones (or are dominated by them in the latter case) achieves regret $\otil(n(\sqrt{kT}+k))$ for rewards in $[0,1]$ and supports of size at most $k$~\cite{ghuge2026semibandit}, where $\otil$ suppresses logarithmic factors.
They also extend their framework to continuous distributions under additional assumptions on the threshold structure of policies and the regularity of the reward functions.

These results leave the following question.
Can the regret grow as $\sqrt T$ without FSD monotonicity, with a polynomial dependence on $n$ that does not involve the number of policies or observation histories, even for continuous distributions?
Treating each deterministic policy as an arm already yields $\sqrt T$ dependence when there are finitely many policies, but the regret then grows with their number.
We answer this question affirmatively, and our results extend the framework of Ghuge, Zhuo, Agarwal, and Nagarajan beyond FSD monotonicity and improve their regret bound.

FSD monotonicity fails in natural problems, and Section~\ref{sec:applications} studies five of them.
In stochastic Boolean function evaluation (SBFE)~\cite{dhk2014}, binary variables are tested at a cost until a known Boolean function is determined, and the expected cost of evaluating the majority function is largest when successes and failures are balanced.
In the full-information secretary problem~\cite{gilbert1966recognizing,esfandiari2020prophets}, a stochastically larger later value makes accepting an earlier candidate less likely to succeed but waiting more likely to succeed.
In optimal stopping with loss aversion~\cite{kleinberg2022optimal}, a large early value raises the reference point and lowers the utility of later candidates.
In scheduling with deadlines~\cite{dean2005thesis} and its special case, correlated stochastic knapsack~\cite{GuptaKMR11}, a larger processing time or size can increase the reward of the current job or item but leaves less room for the others.
In overbooking with group reservations~\cite{subramanian1999}, more arrivals reduce refunds but increase the cost of exceeding capacity.

Under FSD monotonicity, optimistic distributions can be constructed by shifting probability mass in the empirical distributions upward or downward according to the direction of monotonicity.
Without this property, the optimistic distribution can depend on the observation history and need not be confined to a single direction.
The dependence on history already appears in SBFE for the majority of three variables (Figure~\ref{fig:tree-example}): optimism favors increasing the success probability of the next test after observing $1$, but decreasing it after observing $0$, since agreement determines the majority.
The DP for an optimal policy makes this dependence explicit through expectations of \emph{optimal continuation values}, the maximum expected rewards obtainable from each decision point given the observations so far.
Replacing each expectation by its supremum over the corresponding confidence set yields \emph{optimistic continuation values} by backward induction, provided the confidence sets contain the true distributions.

The main challenge is to control the estimation errors in the expectations of these data-dependent values uniformly over all observation histories, with no dependence on the number of histories.
Uniform control over all bounded functions introduces a dependence on the support size and can be vacuous for continuous distributions, so the analysis must exploit the structure of these values.
Moreover, summing variance-insensitive bounds on the local expectation errors yields a regret bound linear in $n$, even when the continuation values are monotone in the newly observed variable.
We address these issues through the level-set structure of the optimistic continuation values and a variance-sensitive analysis of the accumulated expectation errors.


\subsection{Our Contributions}
We refer to the $i$th random variable as coordinate $i$.
For a function of one observed coordinate, its \emph{level-set complexity} is the smallest integer $s$ such that every superlevel set is the intersection of the coordinate's domain with a union of at most $s$ intervals.
Our algorithm replaces each expectation in the DP by its supremum over a confidence set of distributions, i.e., optimistic continuation values.
Our main assumption, which we call the \emph{level-set condition}, bounds by $s_i$ the level-set complexity of the optimistic continuation values after coordinate $i$ is revealed, for any possible confidence sets (Assumption~\ref{ass:level-set}).

\medskip

\noindent
\textbf{Regret Upper Bound.}
With rewards in $[0,1]$, let $k_i$ be the number of support points of the distribution $D_i$ of coordinate $i$, which may be infinite, and let $S=\sum_{i=1}^n\min\{s_i,k_i\}$.
The algorithm's regret satisfies $R_T=\otil(\sqrt{ST})$ against the optimal policy (Theorem~\ref{thm:local}).
This guarantee does not require FSD monotonicity or finite supports, and the algorithm requires no knowledge of $s_i$ or $k_i$.

In particular, when the level-set complexities are bounded by a constant, the regret is $\otil(\sqrt{nT})$, even for continuous distributions and without FSD monotonicity.
If each distribution has at most $k$ support points, the regret is $\otil(\sqrt{nkT})$ even when the supports are unknown and without any condition on the level-set complexity (Corollary~\ref{cor:finite-support}).
In the setting of Ghuge, Zhuo, Agarwal, and Nagarajan, this improves their bound $\otil(n(\sqrt{kT}+k))$ without requiring FSD monotonicity.
For FSD-monotone problems with finite domains, as in their framework, it also suffices to run an exact or $\alpha$-approximation algorithm for known distributions once per round on a single product distribution chosen from the confidence sets (Proposition~\ref{prop:single}), with $\alpha$-regret $\otil(\sqrt{nkT})$ for supports of size at most $k$.
If the problem is realization-monotone, that is, all rewards and all sets of feasible actions are monotone in the observed values in a common direction, then $s_i=1$ (Section~\ref{subsec:regret}), and the regret is $\otil(\sqrt{nT})$.
This applies to the prophet inequality, Pandora's box, and stochastic knapsack with fixed rewards after rescaling the rewards to $[0,1]$, and for the first two with finite domains, each round runs in polynomial time by Proposition~\ref{prop:single}.
The bound $\otil(\sqrt{nT})$ does not depend on the support sizes, whereas the finite-support bound in \cite{ghuge2026semibandit} grows with $k$.
Table~\ref{tab:comparison} compares our general bounds with previous ones.

\medskip

\noindent
\textbf{Regret Lower Bound.}
We complement our upper bounds with matching lower bounds, which show that their dependence on the level-set complexities and the support sizes is necessary.
For any integers $T,s_1,\ldots,s_n\ge1$, we construct a problem satisfying Assumption~\ref{ass:level-set} with these $s_i$ such that, for every learning algorithm, some product distribution with $2s_i+1$ support points for coordinate $i$ forces regret $\Omega(\min\{T,\sqrt{ST}\})$, even under full feedback (Theorem~\ref{thm:lower}).
Here, full feedback means that all $n$ coordinates are observed in each round.
Hence, the worst-case regret over the problems with given $s_1,\ldots,s_n$ is $\thetatil(\min\{T,\sqrt{ST}\})$ with $S=\sum_is_i$.
Since coordinate $i$ of this instance has $2s_i+1$ support points, the dependence on the support sizes in Corollary~\ref{cor:finite-support} cannot be improved in the worst case either.
For realization-monotone problems, our bound $\otil(\sqrt{nT})$ is also tight up to logarithmic factors, since Ghuge, Zhuo, Agarwal, and Nagarajan~\cite{ghuge2026semibandit} observed a lower bound of $\Omega(\sqrt{nT})$ under semi-bandit feedback by a reduction from multi-armed bandits, and their instance with binary coordinates is realization-monotone.


\begin{table}[tb]
\centering\small
\caption{Regret bounds for learning SOO under semi-bandit feedback with rewards in $[0,1]$.
Here, $k_i$ is the support size of $D_i$, $k=\max_ik_i$, and $S=\sum_i\min\{s_i,k_i\}$. We assume $T\ge S$.}
\label{tab:comparison}
\renewcommand{\arraystretch}{1.15}
\begin{tabular}{@{}llll@{}}
\toprule
Result & Condition & Supports & Regret \\
\midrule
\cite{ghuge2026semibandit} & FSD monotonicity & known $k$ points & $\otil\big(n(\sqrt{kT}+k)\big)$ \\
\cite{pmlr-v134-guo21a,ghuge2026semibandit} & none & known $k$ points & $\otil\big((nkT)^{2/3}\big)$ \\
\cite{ghuge2026semibandit} & FSD monotonicity & 2 points & $\Omega(\sqrt{nT})$\\
\midrule
Theorem~\ref{thm:local} & level-set complexity $s_i$ & arbitrary & $\otil(\sqrt{ST})$ \\
Theorem~\ref{thm:local} & realization monotonicity & arbitrary & $\otil(\sqrt{nT})$ \\
Corollary~\ref{cor:finite-support} & none & unknown $k$ points & $\otil(\sqrt{nkT})$ \\
Theorem~\ref{thm:lower} & level-set complexity $s_i$ & $2s_i+1$ points & $\Omega(\sqrt{ST})$\\
\bottomrule
\end{tabular}
\end{table}

\medskip

\noindent
\textbf{Applications.}
For the five applications, we verify the level-set condition directly for SBFE, where $s_i=1$ since the variables are binary, and for the full-information secretary problem and optimal stopping with loss aversion, where $s_i=2$.
Scheduling with deadlines, including correlated stochastic knapsack, and overbooking with group reservations require an extension, since a sum of functions with small level-set complexity need not have small complexity itself.
Theorem~\ref{thm:sum} handles continuation values decomposed into an immediate reward collected when a coordinate is revealed and a residual reward collected later.
It assumes that the immediate rewards have bounded level-set complexity, are nonnegative, and sum to at most 1 in each round, and that larger revealed values cannot increase the residual rewards or enlarge future sets of feasible actions.
Together, the two theorems yield regret $\otil(\sqrt{nT})$ for all five applications after rescaling the rewards to $[0,1]$, provided that, in scheduling, the reward functions have bounded level-set complexity and the observed processing time and reward identify the outcome.
For the full-information secretary problem, the bound holds for every known distribution over arrival orders.

A refinement of Theorem~\ref{thm:local} yields $\otil(\sqrt{ST\,p(1-p)}+S)$, where $p$ is the optimal value (Proposition~\ref{prop:first-order}), and Proposition~\ref{prop:sum-first-order} provides an analogous refinement of Theorem~\ref{thm:sum}.
These refinements yield sharper regret bounds for all five applications in their original units.

\medskip

\noindent
\textbf{Computation Time.}
For computation, we reduce each optimistic update to a linear program (LP) over the masses of the sample points and the intervals between them, which is the dual of a minimum-cost flow problem.
Its coefficients are the suprema of the optimistic continuation value on these parts.
Given these suprema, each round runs in time polynomial in $n$, $t$, and the size of the decision tree (Proposition~\ref{prop:computation}), which can be exponential in $n$, and the suprema themselves can be hard to compute in general.
Each supremum reduces to evaluations of the optimistic continuation value at finitely many points when the domains are finite, or when the domains are $[0,1]$ and the values are continuous and either quasi-convex or monotone on each of finitely many known intervals.
When the values are Lipschitz, approximate evaluations on a grid increase the regret by at most 1 and take time polynomial in $n$, $T$, and the Lipschitz constant.
As in a DP for known distributions, these evaluations can be organized by states when the rewards and the sets of feasible actions depend on the observations only through a state, apart from rewards collected along the way, and each round then runs in time polynomial in $n$, $t$, the size of the tree, and the number of pairs of a node and a state reachable through the evaluation points (Proposition~\ref{prop:states}).
For overbooking with group reservations, this yields time polynomial in $n$, $t$, the number of allowed sets of accepted groups, and the total group size (Appendix~\ref{subsec:states}).
Each round also runs in polynomial time for optimal stopping with loss aversion when each domain is $[0,1]$ or finite (Appendix~\ref{subsec:states}) and for the full-information secretary problem with finite domains and a fixed arrival order (Section~\ref{subsec:secretary}), and it runs in time $O(n^2)$ for SBFE with threshold functions (Section~\ref{subsec:sbfe}).

\subsection{Technical Overview}
Our algorithm follows the principle of optimism in the face of uncertainty.
In our formulation (Section~\ref{sec:problem}), the decision points are the nodes of a finite tree whose edges either reveal a coordinate or stop, and the observed values are carried along the path.
For each coordinate, we construct from the samples collected so far a confidence set that contains the true distribution with high probability, and we replace each expectation in the DP by its supremum over this set.
If the true distributions belong to the confidence sets, backward induction shows that the resulting values are at least the true optimal values.
Each DP update takes the supremum separately, and different nodes and observations may use different distributions from the same confidence set.

The main technical step translates this local optimism into a global regret bound.
The confidence sets impose Bernstein-type constraints on the probabilities of intervals.
For two distributions $P,Q$ in the same confidence set built from $m\ge1$ samples and a $[0,1]$-valued function $\phi$ with level-set complexity at most $s$, Lemma~\ref{lem:level-stability}(b) yields
\[
\left|\E_{X\sim Q}[\phi(X)]-\E_{X\sim P}[\phi(X)]\right|
\le\otil\!\left(
\sqrt{{s}/{m}}\,\sqrt{\Var_{X\sim P}(\phi(X))}+{s}/{m}
\right).
\]
We prove this bound by integrating the probability errors over the superlevel sets of $\phi$, each of which is a union of at most $s$ intervals.
A variance-sensitive inequality for this integral yields the square-root dependence on $\Var_{X\sim P}(\phi(X))$ (Lemma~\ref{lem:layer}).

Fix the history before round $t$ and suppose that the confidence sets contain the true distributions.
Let $Y_{t,h}$ be the optimistic continuation value at the node visited at stage $h$, and let $d_{t,h}=Y_{t,h}-\E[Y_{t,h+1}\mid\FF_{t,h}]$ and $w_{t,h}=\Var(Y_{t,h+1}\mid\FF_{t,h})$, where $\FF_{t,h}$ is the history through stage $h$.
Since $\E[Y_{t,0}]$ is at least the optimal value and the last $Y_{t,h}$ is the reward, the expected regret in this round is at most $\E[\sum_hd_{t,h}]$.
Let $N_{i,t}$ be the number of rounds before round $t$ in which coordinate $i$ was revealed, and let $I_{i,t}$ indicate whether coordinate $i$ is revealed on an edge in round $t$.
Up to logarithmic factors, the proof combines three bounds:
\[
d_{t,h}\lesssim\sqrt{\frac{s_iw_{t,h}}{N_{i,t}+1}}+\frac{s_i}{N_{i,t}+1},
\qquad
\E\biggl[\sum_hw_{t,h}\biggr]\le1,
\qquad
\sum_{t=1}^T\frac{I_{i,t}}{N_{i,t}+1}\le1+\log T.
\]
In the first bound, $i$ is the coordinate revealed at stage $h$.
The first bound follows from the expectation bound above, and it is the only place where level-set complexity is used.
If $D_i$ has $k_i<s_i$ support points, a one-sided version of the expectation bound (Lemma~\ref{lem:finite-support}) yields the first bound with $k_i$ in place of $s_i$, and hence the regret depends on $S=\sum_i\min\{s_i,k_i\}$.
The second bound holds because $(Y_{t,h})_h$ is a supermartingale with values in $[0,1]$, and telescoping the squares $Y_{t,h}^2$ bounds the sum of the conditional variances.
The third bound holds because $N_{i,t}$ increases by one in each round in which coordinate $i$ is revealed.
The Cauchy--Schwarz inequality then yields the leading term $\sqrt{ST}$.

For the lower bound, we construct an instance with $S$ unknown signs $\theta_{ij}$, where $j\in[s_i]$ for coordinate $i$.
Each coordinate $i$ equals a special value $\bot$ with probability $1-s_i/(2S)$.
Otherwise, it takes a value $(j,y)$, where the index $j\in[s_i]$ is uniform and the sign $y\in\{-1,+1\}$ equals $\theta_{ij}$ with probability $(1+\varepsilon)/2$.
Before receiving observations in a round, the learner chooses a prediction $\widehat y_{ij}$ of the observed sign $y$ for each possible pair $(i,j)$.
The reward is 1 if exactly one coordinate differs from $\bot$ and its sign is predicted correctly, and 0 otherwise.
Each prediction $\widehat y_{ij}$ that differs from $\theta_{ij}$ reduces the expected reward by $\Theta(\varepsilon/S)$ relative to predicting $\theta_{ij}$.

Two environments differing in one sign have one-sample Kullback--Leibler (KL) divergence $O(\varepsilon^2/S)$.
Choosing $\varepsilon=\Theta(\min\{1,\sqrt{S/T}\})$ with a sufficiently small constant keeps the probability that $\widehat y_{ij}\ne\theta_{ij}$ bounded below by a constant in every round, averaged over a uniformly random sign vector $\boldsymbol\theta$.
Summing the resulting losses over all $S$ signs and $T$ rounds yields regret $\Omega(\min\{T,\sqrt{ST}\})$ for some fixed sign vector.
In the SOO instance, the policy chooses the prediction vector at the root and then observes all coordinates in order, and we order the $2s_i+1$ values of coordinate $i$ such that every optimistic continuation value has level-set complexity at most $s_i$.

\subsection{Related Work}\label{subsec:related}
For known distributions, prophet inequalities provide guarantees against the expected offline optimum for a single selection~\cite{Krengel1977-uu}, multiple selections~\cite{Alaei14}, and matroid constraints~\cite{kleinberg2012matroid}.
Approximation algorithms are also known for many stochastic combinatorial optimization problems~\cite{DeanGV08,Ma18,asadpour2016maximizing,GN13}.

In combinatorial bandits~\cite{icml/ChenWY13}, the learner selects a set of arms and observes each of them, and algorithms based on upper confidence bounds achieve $\otil(\sqrt{mnT})$ regret for linear rewards when at most $m$ arms are selected~\cite{KvetonWAS15}.
Extensions allow rewards that depend on the whole distributions of the arms~\cite{chen2016general} and additional arms that are revealed depending on the outcomes~\cite{chen2016cmab}.
In our setting, the policy also adapts its actions to the values observed within a round.

SOO can also be cast as an episodic Markov decision process (MDP) whose states are observation histories.
Our algorithm resembles optimistic planning over a set of plausible transition models, as in extended value iteration~\cite{jaksch2010near}, and our variance analysis resembles the use of the law of total variance in the upper confidence bound value iteration algorithm~\cite{AzarOM17}.
However, regret bounds for tabular MDPs depend on the numbers of states and actions, and observation histories can be exponentially many in $n$, or infinitely many for continuous observations.
Factored MDPs~\cite{tian2020factored} exploit a product structure of the transitions, but their bounds still depend on the sizes of the factored state spaces.
In SOO, all histories share the same coordinate distributions, and the level-set condition on the optimistic continuation values removes the dependence on the numbers of histories and support points.

Other work on learning in SOO considers specific problems.
Gatmiry, Kesselheim, Singla, and Wang~\cite{Gatmiry2024} proved an $\otil(n^3\sqrt T)$ upper bound and an $\Omega(\sqrt T)$ lower bound for the prophet inequality under bandit feedback.
Liu, Chen, Wang, Luo, and Ratliff~\cite{liu2025improved} obtained $\otil(\sqrt{nT})$ for the prophet inequality and Pandora's box under semi-bandit feedback by using problem-specific structure.
For Pandora's box with inspection costs $c_i$ and values in $[0,1]$, their bound avoids the factor $1+\sum_i c_i$ introduced by normalizing the net reward in our general bound.

A related line of work studies the sample complexity of stochastic optimization, with bounds for single-parameter revenue maximization~\cite{guo2019settling}, the prophet inequality~\cite{correa2019prophet}, and general problems over product distributions~\cite{pmlr-v134-guo21a}.
For the last class, Guo, Huang, Tang, and Zhang~\cite{pmlr-v134-guo21a} show that $\otil(nk/\varepsilon^2)$ samples suffice for supports of size at most $k$.
Under a stronger monotonicity condition, $\otil(n/\varepsilon^2)$ samples suffice, and both bounds are optimal up to logarithmic factors.
Under full feedback, such bounds lead to regret with $\sqrt T$ dependence, whereas under semi-bandit feedback, an explore-then-commit approach typically leads to $T^{2/3}$ dependence~\cite{ghuge2026semibandit}.

In concurrent and independent work that appeared on arXiv shortly before this submission, Zhuo and Nagarajan~\cite{zhuo2026nearoptimal} and Hu and Ghuge~\cite{hg2026} improved the regret bounds for FSD-monotone problems.

\section{Preliminaries}\label{sec:problem}
\subsection{Model}
Let $\mathbb Z_{>0}$ be the set of positive integers.
For $m\in\mathbb Z_{>0}$, write $[m]=\{1,\ldots,m\}$, and for $a\in\mathbb R$, write $a_+=\max\{a,0\}$.
Let $X_1,\ldots,X_n$ be independent random variables with unknown distributions $D_1,\ldots,D_n$, and write $D=\prod_{i=1}^nD_i$.
Each $X_i$ takes values in a known subset of $[0,1]$, called its domain (e.g., $\{0,1\}$ or $[0,1]$).
Our algorithm and analysis use only the order of the values, and hence real-valued coordinates can be mapped into $[0,1]$ by a known strictly increasing function.
We refer to $X_i$ as coordinate $i$ and say that coordinate $i$ is revealed when its realized value is observed.

A problem is specified by a finite rooted directed tree $(V,A)$, called the decision tree,\footnote{The analysis applies, with subtrees replaced by the subgraphs reachable from a node, to a finite directed acyclic graph with a single root in which all paths from the root to each node reveal the same coordinates, and hence $\obs(u)$ below is well defined.
Such a graph can be much smaller than the corresponding tree, as in SBFE, where a node can represent the set of tested coordinates.
We use trees for simplicity.} with root $r$ and children $\child(u)$ at node $u$, together with the observation sets, feasible actions, and leaf rewards defined below.
The set $\obs(u)\subseteq[n]$ contains the coordinates already observed upon reaching $u$, and $x_{\obs(u)}=(x_i)_{i\in\obs(u)}$ denotes their values.
The coordinates in $\obs(r)$ are observed before the first action.
Each edge $(u,v)$ is either a \emph{revelation edge}, which reveals one new coordinate $i$, with $\obs(v)=\obs(u)\cup\{i\}$ and $i\notin\obs(u)$, or a \emph{stopping edge}, which leads to a leaf $v$ without a new observation, with $\obs(v)=\obs(u)$.
Thus, a coordinate is observed at most once per round.
We remark that an action that reveals several coordinates at once can be represented by a chain of revelation edges with no intermediate choices.
A revelation edge may also lead to a leaf. 

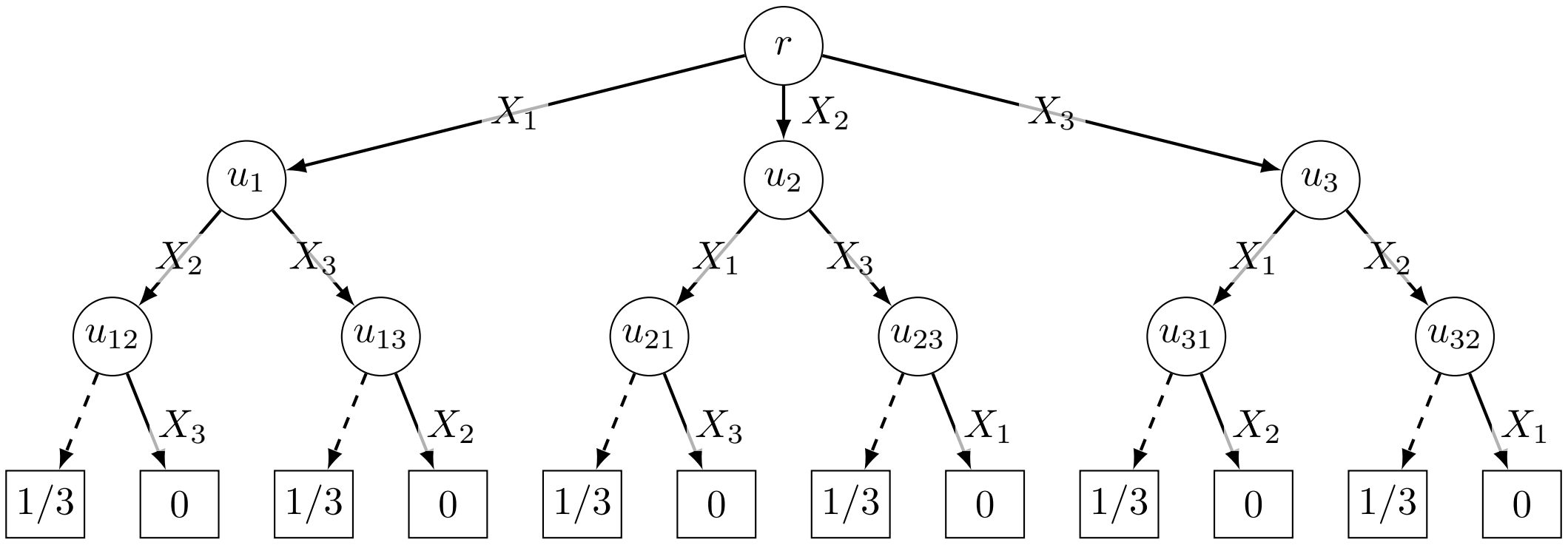
\begin{figure}[htb]
\centering
\begin{tikzpicture}[
  font=\small,
  inner/.style={circle,draw,minimum size=7mm,inner sep=0pt},
  leaf/.style={rectangle,draw,minimum width=7mm,minimum height=6mm,inner sep=1pt},
  reveal/.style={-latex,thick},
  stop/.style={-latex,thick,dashed},
  lab/.style={inner sep=2pt,fill=white,fill opacity=0.7,text opacity=1},
]
\node[inner] (r) at (0,0) {$r$};
\foreach \i/\pos in {1/-4.8,2/0,3/4.8} {
  \node[inner] (u\i) at (\pos,-1.2) {$u_\i$};
  \ifnum\i=2
    \draw[reveal] (r) -- node[lab,midway,right=2pt] {$X_\i$} (u\i);
  \else
    \draw[reveal] (r) -- node[lab,midway] {$X_\i$} (u\i);
  \fi
}
\foreach \i/\j/\k/\pos in {1/2/3/-6,1/3/2/-3.6,2/1/3/-1.2,2/3/1/1.2,3/1/2/3.6,3/2/1/6} {
  \node[inner] (u\i\j) at (\pos,-2.6) {$u_{\i\j}$};
  \node[leaf] (s\i\j) at (\pos-0.6,-4.1) {$1/3$};
  \node[leaf] (l\i\j\k) at (\pos+0.6,-4.1) {$0$};
  \draw[reveal] (u\i) -- node[lab,midway] {$X_\j$} (u\i\j);
  \draw[stop] (u\i\j) -- (s\i\j);
  \draw[reveal] (u\i\j) -- node[lab,right,pos=0.55] {$X_\k$} (l\i\j\k);
}
\end{tikzpicture}
\caption{SBFE (Section~\ref{subsec:sbfe}) for the majority function of three variables, $F(x)=\ind\{x_1+x_2+x_3\ge2\}$, with unit testing costs.
Here, $\obs(r)=\emptyset$, $\obs(u_i)=\{i\}$, and $\obs(u_{ij})=\{i,j\}$.
At $u_{ij}$, the dashed stopping edge is feasible if $x_i=x_j$, and the solid edge revealing the remaining coordinate is feasible if $x_i\ne x_j$.
Each leaf shows its reward, 1 minus the number of tests divided by 3.}
\label{fig:tree-example}
\end{figure}

We regard each child of a node as an action at that node, and at a non-leaf node $u$ with observations $x_{\obs(u)}$, the \emph{feasible actions} form a nonempty set $\mathcal A_u(x_{\obs(u)})\subseteq\child(u)$.
A leaf $\ell$ has reward $f_\ell(x_{\obs(\ell)})\in[0,1]$.
The tree encodes only the actions and the order of observations, and the observed values are carried along as $x_{\obs(u)}$.
In the example of Figure~\ref{fig:tree-example}, a single node $u_{12}$ serves all pairs $(x_1,x_2)$, and its feasible actions depend on whether $x_1=x_2$.
Reward functions and functions specifying feasibility are assumed to be measurable.

A deterministic policy $\sigma$ chooses a feasible action $\sigma_u(x_{\obs(u)})\in\mathcal A_u(x_{\obs(u)})$ at each non-leaf node $u$.
Deterministic policies are assumed to be measurable, and a randomized policy is a distribution over them.
For a policy $\sigma$ and a realization $x$, write $f(\sigma,x)$ for the reward, averaged over the randomness of $\sigma$, and let
\[
f(\sigma\mid D)=\E_{X\sim D}[f(\sigma,X)],
\qquad \opt(D)=\sup_\sigma f(\sigma\mid D).
\]

The learner faces the same problem in each of $T$ rounds.
Before the first round, the learner knows the problem and $T$, but not $D$.
In each round $t\in[T]$, a vector $X^t\sim D$ is drawn independently of the previous rounds.
The learner chooses a policy $\sigma^t$ based on the observations from the previous rounds and possibly internal randomness, and $\sigma^t$ adapts to the observations within round $t$.
If $\ell^t$ is the leaf reached, the learner observes $X^t_{\obs(\ell^t)}$, which we call semi-bandit feedback.
Let $N_{i,t}$ be the number of rounds before round $t$ in which coordinate $i$ was revealed, and let $Z_{i,1},\ldots,Z_{i,N_{i,t}}$ be the revealed values in order.

The expected regret of the policy $\sigma^t$ chosen in round $t$ is $\reg_t=\opt(D)-f(\sigma^t\mid D)$.
Taking expectation over the past observations and the learner's randomness, we define the cumulative expected regret as
\begin{equation*}
R_T=\sum_{t=1}^T\E[\reg_t]
=\sum_{t=1}^T\bigl(\opt(D)-\E[f(\sigma^t,X^t)]\bigr).
\end{equation*}
We write $R_T(D)$ when the dependence on $D$ matters.
The benchmark $\opt(D)$ is attained by a policy that knows $D$ but, like the learner, observes the coordinates only along its execution path.
For $\alpha\in(0,1]$, we also define the cumulative expected
$\alpha$-regret as
\[
R_T^{(\alpha)}
=\sum_{t=1}^T
\bigl(\alpha\,\opt(D)-\E[f(\sigma^t,X^t)]\bigr).
\]

We define functions $g_u$ and $h_{(u,v)}$ by backward induction.
At a leaf, $g_\ell=f_\ell$.
At a non-leaf node,
\begin{align}
g_u(x_{\obs(u)})&=\textstyle\max_{v\in\mathcal A_u(x_{\obs(u)})}h_{(u,v)}(x_{\obs(u)}),\label{eq:true-dp}\\
h_{(u,v)}(x_{\obs(u)})&=
\begin{cases}
\E_{X_i\sim D_i}[g_v(X_i,x_{\obs(u)})],&\text{if }\obs(v)\setminus\obs(u)=\{i\},\\
f_v(x_{\obs(u)}),&\text{if }\obs(v)=\obs(u).
\end{cases}\label{eq:true-edge}
\end{align}
Here, $(x_i,x_{\obs(u)})$ adds the value of coordinate $i$ to the observations, with coordinates identified by their indices.
We call $g_u(x_{\obs(u)})$ the optimal continuation value at $u$.
It is the maximum expected reward obtainable after reaching $u$ with observations $x_{\obs(u)}$, and $h_{(u,v)}(x_{\obs(u)})$ is the corresponding maximum when the next move is to $v$.
By backward induction, choosing a maximizing child at every node is optimal, and
\begin{equation}\label{eq:opt}
\opt(D)=\E_{X_{\obs(r)}\sim\prod_{i\in\obs(r)}D_i}[g_r(X_{\obs(r)})].
\end{equation}

We now relate our model to that of Ghuge, Zhuo, Agarwal, and Nagarajan~\cite{ghuge2026semibandit}.
They describe a policy as a tree whose internal nodes are labeled by the coordinates to observe next and whose branches correspond to the observed values.
When the constraints on policies are imposed on each path, such a problem fits our model, with the nodes of $(V,A)$ corresponding to the sequences of observed coordinates.
For continuous distributions, they consider policies with finitely many threshold branches at each node, since an arbitrary policy can have uncountably many branches.
In our applications, an optimal policy can depend on the exact observed values, for example, through the maximum of the past values in optimal stopping with loss aversion, and it may then need uncountably many branches in their model.
Our model carries the observed values along a finite tree, and a policy chooses a child at each node by a measurable function of these values.
Carrying the values also makes each continuation value, such as $g_v(\cdot,x_{\obs(u)})$, a function of the coordinate revealed on the edge $(u,v)$, and our regret bounds depend on the level-set complexity of such functions (Section~\ref{sec:algorithm}).

\subsection{Monotonicity and Level-Set Complexity}\label{subsec:level-set}
For distributions $P,Q$, we say that $P$ \emph{first-order stochastically dominates} (FSD) $Q$ if $P([0,x])\le Q([0,x])$ for all $x\in[0,1]$.
A problem is \emph{FSD-monotone}~\cite{ghuge2026semibandit} if either $\opt(D)\le\opt(E)$ for all product distributions $D=\prod_iD_i$ and $E=\prod_iE_i$ such that $E_i$ dominates $D_i$ for every $i$, or $\opt(D)\ge\opt(E)$ for all such $D$ and $E$.
A problem is \emph{realization-monotone} if the leaf rewards and the sets of feasible actions are all nondecreasing in every observed value, or all nonincreasing in every observed value, with the other values fixed and sets ordered by inclusion.
If, for each coordinate $i$, the leaf rewards and the sets of feasible actions are monotone in $x_i$ in a direction that depends on $i$, reversing the order of the values of the coordinates with the nonincreasing direction, which affects neither the algorithm nor level-set complexity, makes the problem realization-monotone.
Realization monotonicity implies FSD monotonicity, since expectations and maxima over sets of feasible actions that are monotone in the same direction preserve monotonicity, and hence backward induction shows that every continuation value is monotone in each observed value and that the optimal value is monotone in each distribution.
The converse fails, since FSD monotonicity concerns only the optimal value.
For example, if the root has a stopping edge to a leaf with reward 1 and an edge revealing $X_1$ to a leaf with reward $\psi(x_1)\in[0,1]$, the optimal value is always 1, whereas $\psi$ need not be monotone and can have any level-set complexity.

For an integer $s\ge1$, let $\II_s$ be the family of unions of at most $s$ intervals in $[0,1]$, including the empty set.
Intervals may be open, closed, half-open, or singletons.
A function $\phi$ on a subset of $[0,1]$ has \emph{level-set complexity at most $s$} if, for every $y\in\mathbb R$, there exists $B\in\II_s$ such that the \emph{superlevel set} of $\phi$ at $y$ satisfies
\[
\{z\in\operatorname{dom}(\phi):\phi(z)>y\}=\operatorname{dom}(\phi)\cap B.
\]
By convention, every function has level-set complexity at most $\infty$.

A function has level-set complexity at most 1 if it is monotone or unimodal, that is, nondecreasing and then nonincreasing.
A function has level-set complexity at most 2 if it is nonincreasing and then nondecreasing, for example, the maximum of a nondecreasing function and a nonincreasing function, since each superlevel set is then the intersection of the domain with the complement of an interval.
When the domain is an interval, the same bounds hold for quasi-concave and quasi-convex functions, respectively.
A function on a domain of at most $k$ points has level-set complexity at most $\lceil k/2\rceil$, since each superlevel set consists of at most that many runs of consecutive domain points.

\section{Main Results}\label{sec:algorithm}
We present an optimistic DP algorithm and state regret bounds in terms of level-set complexity, together with a worst-case lower bound that matches the main upper bound up to logarithmic factors.
We then extend the upper bound to problems whose reward decomposes into the reward earned so far and the future reward, such as scheduling and overbooking.

\subsection{Optimistic Dynamic Programming}\label{subsec:optimistic-dp}
At the beginning of each round, the algorithm forms a confidence set of distributions for each coordinate from the samples collected so far.
It then evaluates the DP defined by~\eqref{eq:true-dp} and~\eqref{eq:true-edge} with each expectation replaced by its supremum over the corresponding confidence set, and during the round it chooses, at each node, a child that maximizes the resulting optimistic value.

For each $m\ge1$, write $\widehat D_{i,m}$ for the empirical distribution of the first $m$ revealed values $Z_{i,1},\ldots,Z_{i,m}$ of coordinate $i$, that is, $\widehat D_{i,m}(B)=\frac1m\sum_{j=1}^m\ind\{Z_{i,j}\in B\}$ for $B\subseteq[0,1]$.
For $\delta\in(0,1)$, let
\begin{equation*}
\Lambda=\log\left({80n(T+1)^3}/{\delta}\right).
\end{equation*}
For $m\ge1$ and $z\in[0,1]$, define
\begin{equation}\label{eq:radius}
b_m(z)=\min\left\{1,\;6\sqrt{z(1-z)\Lambda/m}+32\Lambda/m\right\}.
\end{equation}
The confidence set for $m\ge1$ observations of coordinate $i$ is
\begin{equation}\label{eq:confidence}
\CC_m(\widehat D_{i,m})=
\left\{Q:
|Q(I)-\widehat D_{i,m}(I)|\le b_m(\widehat D_{i,m}(I))
\quad\forall I\in\II_1\right\},
\end{equation}
where $Q$ ranges over distributions on the coordinate's domain.
For $m=0$, let $\widehat D_{i,0}$ be an arbitrary fixed distribution on the coordinate's domain and let $\CC_0(\widehat D_{i,0})$ be the set of all distributions on that domain.
For $m\ge1$, the empirical distribution belongs to the confidence set, and hence the set is always nonempty.
For $m\le32\Lambda$, we have $b_m\equiv1$, and the confidence set consists of all distributions on the coordinate's domain.

At the beginning of round $t$, set $\CC_{i,t}=\CC_{N_{i,t}}(\widehat D_{i,N_{i,t}})$.
These sets stay fixed throughout the round.
Define $\bar g_\ell^{\,t}=f_\ell$ at leaves and
\begin{align}
\bar g_u^{\,t}(x_{\obs(u)})&=\max\nolimits_{v\in\mathcal A_u(x_{\obs(u)})}\bar h_{(u,v)}^{\,t}(x_{\obs(u)}),
\label{eq:optimistic-dp}\\
\bar h_{(u,v)}^{\,t}(x_{\obs(u)})&=
\begin{cases}
\sup_{Q\in\CC_{i,t}}\E_{X_i\sim Q}[\bar g_v^{\,t}(X_i,x_{\obs(u)})],
&\text{if }\obs(v)\setminus\obs(u)=\{i\},\\
f_v(x_{\obs(u)}),&\text{if }\obs(v)=\obs(u).
\end{cases}\label{eq:optimistic-edge}
\end{align}
We call $\bar g_u^{\,t}(x_{\obs(u)})$ the optimistic continuation value at $u$.
The suprema at different edges may be attained by different distributions, and no single product distribution needs to attain all optimistic values.
Upon reaching node $u$, the algorithm chooses
\begin{equation}\label{eq:policy-choice}
\textstyle\sigma_u^t(x_{\obs(u)})\in\mathop{\mathrm{arg\,max}}_{v\in\mathcal A_u(x_{\obs(u)})}
\bar h_{(u,v)}^{\,t}(x_{\obs(u)}),
\end{equation}
breaking ties in a fixed order.
Appendix~\ref{app:lp} shows that each supremum in~\eqref{eq:optimistic-edge} is the optimal value of an LP whose size is polynomial in the number of samples, given the suprema of $\bar g_v^{\,t}$ over the cells formed by the sample points and the intervals between them.
For infinite domains, we assume that the optimistic values are measurable in the samples and observations, which makes this policy measurable.
This holds automatically for finite domains.
Algorithm~\ref{alg:main} summarizes the procedure.

\begin{algorithm}[htb]
\caption{Learning with optimistic dynamic programming}\label{alg:main}
\KwIn{parameter $\delta\in(0,1)$}
\For{$t=1,\ldots,T$}{
Set $\CC_{i,t}=\CC_{N_{i,t}}(\widehat D_{i,N_{i,t}})$ for every coordinate $i$, as defined above\;
Define $\bar g^{\,t}$ and $\bar h^{\,t}$ by~\eqref{eq:optimistic-dp} and~\eqref{eq:optimistic-edge}\;
Observe $x_{\obs(r)}$ and set $u\gets r$\;
\While{$u$ is not a leaf}{
Choose a child $v$ of $u$ by~\eqref{eq:policy-choice}\;
Observe any coordinate revealed by $(u,v)$ and set $u\gets v$\;
}
Add the values revealed in round $t$ to the samples\;
}
\end{algorithm}

\subsection{Regret Bounds}\label{subsec:regret}
The regret depends on how the optimistic continuation values vary with the coordinate revealed next.
On an edge $(u,v)$ that reveals coordinate $i$, the optimistic value $\bar h_{(u,v)}^{\,t}$ is the supremum of the expectation of $z\mapsto\bar g_v^{\,t}(z,x_{\obs(u)})$ over $\CC_{i,t}$, as defined in~\eqref{eq:optimistic-edge}.
We bound the difference between this optimistic value and the expectation of the same continuation-value function under $D_i$.
Since expectations can be expressed as integrals of superlevel-set probabilities, the interval constraints control this difference whenever $D_i\in\CC_{i,t}$ and each superlevel set is a union of a few intervals.
We assume the following bound on the level-set complexity of this function for any confidence sets.
\begin{assumption}\label{ass:level-set}
There are $s_1,\ldots,s_n\in\mathbb Z_{>0}\cup\{\infty\}$ with the following property.
Let $\PP_1,\ldots,\PP_n$ be any nonempty sets of distributions on the domains of the coordinates, and let $\bar g$ be defined by~\eqref{eq:optimistic-dp} and~\eqref{eq:optimistic-edge} with $\PP_i$ in place of $\CC_{i,t}$.
Then, for every edge $(u,v)$ revealing coordinate $i$ and all observations $x_{\obs(u)}$ for which $v\in\mathcal A_u(x_{\obs(u)})$, the function $z\mapsto\bar g_v(z,x_{\obs(u)})$ has level-set complexity at most $s_i$.
\end{assumption}
We can take $s_i=1$ for every coordinate if the problem is realization-monotone, since the backward induction in Section~\ref{subsec:level-set}, which also applies to suprema, shows that every optimistic continuation value is then monotone.
These parameters are small constants in many problems, including SBFE, the full-information secretary problem, and optimal stopping with loss aversion in Section~\ref{sec:applications}.
If the domain of coordinate $i$ is finite, the assumption holds with $s_i$ equal to half its size rounded up, which is 1 for binary coordinates.
The assumption always holds with $s_i=\infty$, for which it imposes no condition on coordinate $i$.

The following theorem bounds the regret in terms of these complexities and the support sizes.
\begin{theorem}\label{thm:local}
Under Assumption~\ref{ass:level-set}, let $k_i\in\mathbb Z_{>0}\cup\{\infty\}$ be the number of support points of $D_i$, and suppose that $S=\sum_{i=1}^n\min\{s_i,k_i\}$ is finite.
There is a universal constant $c$ such that, for every $\delta\in(0,1)$, Algorithm~\ref{alg:main} satisfies
\begin{equation}\label{eq:local-theorem}
R_T\le c\bigl(\sqrt{ST\Lambda\log T(1+\log T)}+S\Lambda(1+\log T)\bigr)+\delta T.
\end{equation}
In particular, for $\delta=1/(T+1)$, we have $R_T=\otil(\sqrt{ST})$.
\end{theorem}
The last statement holds because $\delta T<1$ and $S\le\sqrt{ST}$ when $T\ge S$, whereas $R_T\le T\le\sqrt{ST}$ when $T<S$.
Algorithm~\ref{alg:main} requires no knowledge of $s_i$ or $k_i$, adapts to the smaller of the two for each coordinate, and uses $T$ only through $\Lambda$.
For unknown $T$, the doubling trick, restarting the algorithm in phases of lengths $H=2^j$ for $j=0,1,\ldots$ with $H$ in place of $T$ and $\delta=1/(H+1)$, yields $R_T=\otil(\sqrt{ST})$ for every $T$.

Since Assumption~\ref{ass:level-set} holds with $s_i=\infty$, Theorem~\ref{thm:local} yields the following corollary, which bounds the regret in terms of the support sizes even when the supports are unknown.
\begin{corollary}\label{cor:finite-support}
Suppose that each distribution $D_i$ has at most $k_i$ support points, and let $K=k_1+\cdots+k_n$.
Then, for $\delta=1/(T+1)$, Algorithm~\ref{alg:main} satisfies $R_T=\otil(\sqrt{KT})$, without any condition on the level-set complexity.
\end{corollary}
If the supports were known, we could restrict the coordinate domains to them and apply Theorem~\ref{thm:local} with $s_i=\lceil k_i/2\rceil$.
Corollary~\ref{cor:finite-support} requires neither the supports nor their sizes as input and allows infinite coordinate domains.
Restricting the continuation-value functions to the true supports does not suffice, since distributions in the confidence sets may place mass outside these supports, and the proof of Theorem~\ref{thm:local} accounts for this mass through Lemma~\ref{lem:finite-support}.

The analysis also applies to algorithms that choose a single product distribution from the confidence sets, provided that its optimal value is at least the true one up to $1/T$.
For a product distribution $P=\prod_iP_i$, let $g^P$ and $h^P$ be defined by~\eqref{eq:true-dp} and~\eqref{eq:true-edge} with $P_i$ in place of $D_i$.
A policy is \emph{optimal for $P$ at every node} if, at every non-leaf node and for all observations, it chooses a child that maximizes $h^P$.
\begin{proposition}\label{prop:single}
Suppose $\obs(r)=\emptyset$ and let $\delta=1/(T+1)$.
For each round $t$, let $Q^t=\prod_iQ_i^t$ be a product distribution
determined by the observations before round $t$, with
$Q_i^t\in\CC_{i,t}$ for every $i$, and satisfying
$\opt(Q^t)\ge\opt(D)-1/T$ whenever $D_i\in\CC_{i,t}$ for every $i$.
\begin{enumerate}[label=(\alph*),nosep]
\item Suppose that, for every product distribution $P$, every edge
$(u,v)$ revealing a coordinate $i$, and all observations
$x_{\obs(u)}$ for which $v\in\mathcal A_u(x_{\obs(u)})$,
the function $z\mapsto g_v^P(z,x_{\obs(u)})$ has level-set
complexity at most $s_i$, and let $S$ be as in
Theorem~\ref{thm:local}.
If the algorithm runs a policy that is optimal for $Q^t$ at every node,
then $R_T=\otil(\sqrt{ST})$.
\item Let $\alpha\in(0,1]$, suppose that each $D_i$ has at most
$k_i$ support points, and let $K=k_1+\cdots+k_n$.
If the algorithm runs a policy $\sigma^t$ with
$f(\sigma^t\mid Q^t)\ge\alpha\,\opt(Q^t)$, then
$R_T^{(\alpha)}\le\otil(\sqrt{KT})$.
\end{enumerate}
\end{proposition}

The condition $\obs(r)=\emptyset$ can be satisfied by adding
dummy nodes before the original root and revealing the initial
observations along the resulting chain of edges.
Thus, it does not restrict the SOO problems that can be represented.

A distribution $Q^t$ that satisfies the conditions in the proposition always exists, although finding
it can be difficult in general.
For example, it can be computed when every domain is finite
and the optimal value is nondecreasing under FSD.
For each $i$, we can solve an LP to find a distribution
$Q_i^t\in\CC_{i,t}$ that stochastically dominates every distribution
in $\CC_{i,t}$ (Lemma~\ref{lem:dominating}).
Then $Q^t=\prod_iQ_i^t$ satisfies $\opt(Q^t)\ge\opt(D)$
whenever $D_i\in\CC_{i,t}$ for every $i$.

Part~(b) shows that, if an $\alpha$-approximation policy is known
for the underlying SOO problem with known distributions, we can
incorporate it into this approach, as in the algorithm of
Ghuge, Zhuo, Agarwal, and Nagarajan~\cite{ghuge2026semibandit}.
The resulting $\alpha$-regret is $\otil(\sqrt{KT})$.

The variance bound in the proof of Theorem~\ref{thm:local} can be refined to depend on the optimal value.
\begin{proposition}\label{prop:first-order}
Under Assumption~\ref{ass:level-set}, let $S$ be as in Theorem~\ref{thm:local} and $p=\opt(D)$.
For $\delta=1/(T+1)$, Algorithm~\ref{alg:main} satisfies $R_T=\otil(\sqrt{ST\,p(1-p)}+S)$.
\end{proposition}
This bound is smaller than that of Theorem~\ref{thm:local} when the optimal value is close to 0 or 1.
Section~\ref{subsec:main-upper} proves Theorem~\ref{thm:local} and Propositions~\ref{prop:single} and~\ref{prop:first-order}, using the confidence-set bounds from Section~\ref{subsec:coverage}.

The next theorem shows that, up to logarithmic factors, the dependence on $S$ and $T$ in Theorem~\ref{thm:local} is necessary, even under full feedback.

\begin{theorem}\label{thm:lower}
For any integers $T\ge1$ and $s_1,\ldots,s_n\ge1$, there is a problem with $n$ coordinates that has the following properties.
\begin{enumerate}[label=(\roman*),nosep]
\item Every coordinate has a finite domain, and every leaf reward is 0 or 1.
The learner receives full feedback, that is, all $n$ coordinates are observed in every round under every policy.
\item Assumption~\ref{ass:level-set} holds with $s_1,\ldots,s_n$.
\item For every randomized learning algorithm, there is a product distribution $D$ for which $R_T(D)\ge\frac1{64}\min\{T,\sqrt{ST}\}$, where $S=s_1+\cdots+s_n$ and each $D_i$ has $2s_i+1$ support points.
\end{enumerate}
\end{theorem}
The proof is in Section~\ref{sec:lower}.
Together with Theorem~\ref{thm:local}, it shows that the worst-case regret over the problems satisfying Assumption~\ref{ass:level-set} with $s_1,\ldots,s_n$ is $\thetatil(\min\{T,\sqrt{ST}\})$ for $S=\sum_is_i$.
Since coordinate $i$ of this instance has $2s_i+1$ support points, the bound of Corollary~\ref{cor:finite-support} is also tight up to logarithmic factors in the worst case over supports of these sizes.
The lower-bound instance has $2^S$ root actions, and the rate $\sqrt{ST}$ coincides with the rate $\sqrt{T\log2^S}$ of prediction with expert advice over $2^S$ actions under full feedback~\cite{cesa2006prediction}.
The number of actions does not affect the upper bound, since Theorem~\ref{thm:local} does not depend on the size of the tree.

\subsection{Extension to Immediate Rewards}\label{subsec:extension}
In some applications, the continuation value decomposes into an immediate reward and a residual value.
Their sum can have large level-set complexity even when each term has small complexity.
For example, in overbooking with group reservations, the refund savings of a group are nondecreasing and the remaining net revenue is nonincreasing in its show-up count, but the optimistic continuation value can alternate between two values as this count increases (Section~\ref{subsec:overbooking}).
The following theorem bounds the regret by controlling the expectation errors of the two terms separately.
We assume that the immediate rewards are nonnegative and sum to at most 1 along each root-to-leaf path consisting of feasible actions.
We also require that increasing the revealed value, with all other values fixed, cannot increase the residual reward or enlarge the set of feasible actions at any subsequent node.
\begin{theorem}\label{thm:sum}
Let $s_1,\ldots,s_n\ge1$ be integers and $S=\sum_i s_i$.
Suppose that each revelation edge $e=(u,v)$ has a measurable immediate reward $\chi_e(x_{\obs(v)})\in[0,1]$ satisfying the following conditions.
\begin{enumerate}[label=(\alph*),nosep]
\item For every edge $e=(u,v)$ revealing $i$, fix any $x_{\obs(u)}$ for which $v\in\mathcal A_u(x_{\obs(u)})$.
The function $\chi_e(\cdot,x_{\obs(u)})$ has level-set complexity at most $s_i$, and every leaf $\ell$ in the subtree rooted at $v$ admits the decomposition
\[
f_\ell(x_{\obs(\ell)})=\kappa_e(x_{\obs(u)})+\chi_e(x_{\obs(v)})+F_{e,\ell}(x_{\obs(\ell)}),
\qquad 0\le F_{e,\ell}\le1,
\]
where $\kappa_e$ is common to all such leaves and the identity holds for all remaining coordinate values.
\item For each edge and fixed observations in~(a), the residual rewards $F_{e,\ell}$ and the sets of feasible actions in the subtree rooted at $v$ are nonincreasing in $x_i$, with other coordinates fixed and sets ordered by inclusion.
\item The immediate rewards sum to at most 1 along every root-to-leaf path consisting of feasible actions.
\end{enumerate}
Then, for every $\delta\in(0,1)$, Algorithm~\ref{alg:main} satisfies~\eqref{eq:local-theorem} with a larger universal constant $c$.
In particular, for $\delta=1/(T+1)$, we have $R_T=\otil(\sqrt{ST})$.
\end{theorem}
Since the supremum of a sum is at most the sum of the suprema, the proof bounds the expectation errors separately for the immediate reward and the residual continuation value, and hence no level-set bound on their sum is needed.
As in Theorem~\ref{thm:local}, the bound also holds with $S=\sum_i\min\{s_i,k_i\}$, since Lemma~\ref{lem:finite-support} applies to the immediate rewards if $k_i<s_i$.

When the total immediate reward is bounded by the leaf reward, the regret bound can be refined in terms of the optimal value.
\begin{proposition}\label{prop:sum-first-order}
In addition to the conditions of Theorem~\ref{thm:sum}, suppose that the sum in~(c) is at most the reward of the leaf at the end of the path, and let $p=\opt(D)$.
For $\delta=1/(T+1)$, Algorithm~\ref{alg:main} satisfies $R_T=\otil(\sqrt{STp}+S)$.
\end{proposition}
Section~\ref{subsec:sums} proves Theorem~\ref{thm:sum} and Proposition~\ref{prop:sum-first-order}.

\section{Applications}\label{sec:applications}
We apply Theorems~\ref{thm:local} and~\ref{thm:sum} to five problems, none of which is FSD-monotone in general.
In each problem, the objective lies in a known interval of length $\Gamma>0$.
We rescale it to a reward in $[0,1]$ by an affine map.
For cost minimization, the reward is 1 minus the incurred cost divided by $\Gamma$.
Regret in the original objective is then $\Gamma$ times the normalized regret.
Throughout this section, we take $\delta=1/(T+1)$.
When the scheduling reward functions have bounded level-set complexity, all five problems admit regret $\otil(\Gamma\sqrt{nT})$ in the original objective, and Table~\ref{tab:applications} lists sharper bounds that depend on the optimal values.

\begin{table}[htb]
\centering\small
\caption{Applications in Section~\ref{sec:applications}, with regret in the original objective up to logarithmic factors.
Here, $\Gamma$ is the normalizing constant, $C^*$, $p^*$, $U^*$, $\opt^*$, and $W^*$ are the optimal values, $C$ is the largest overflow cost, and the scheduling reward functions $r_i$ are unimodal.
For scheduling and overbooking, $s_i$ is the parameter of Theorem~\ref{thm:sum}.}
\label{tab:applications}
\renewcommand{\arraystretch}{1.2}
\begin{tabularx}{\textwidth}{@{}l>{\raggedright\arraybackslash}Xccl@{}}
\toprule
Problem & Why FSD monotonicity fails & $s_i$ & Reference & Regret \\
\midrule
SBFE & balanced outcomes need the most tests & 1 & Prop.~\ref{prop:first-order} & $\sqrt{nT\,C^*(\Gamma-C^*)}+\Gamma n$ \\
Secretary & larger later values favor waiting & 2 & Prop.~\ref{prop:first-order} & $\sqrt{nT\,p^*(1-p^*)}+n$ \\
Loss aversion & early values raise the reference point & 2 & Prop.~\ref{prop:first-order} & $\sqrt{nT\,(U^*+\lambda)(1-U^*)}+\Gamma n$ \\
Scheduling & longer jobs delay later jobs & 1 & Prop.~\ref{prop:sum-first-order} & $\sqrt{nT\,\Gamma\opt^*}+\Gamma n$ \\
Overbooking & arrivals save refunds but cause overflow & 1 & Prop.~\ref{prop:sum-first-order} & $\sqrt{nT\,\Gamma(W^*+C)}+\Gamma n$ \\
\bottomrule
\end{tabularx}
\end{table}

\subsection{Stochastic Boolean Function Evaluation}\label{subsec:sbfe}
In SBFE, a known Boolean function $F:\{0,1\}^n\to\{0,1\}$ and nonnegative testing costs $c_1,\ldots,c_n$ are given, and the success probabilities of the independent binary variables $X_i$ are unknown.
A policy tests coordinates one at a time until the observed values determine $F(X)$, that is, until all completions of the unobserved coordinates yield the same function value.
The objective is to minimize the expected testing cost.
SBFE is a basic model of sequential testing, and approximation algorithms for known probabilities have been studied~\cite{dhk2014}.

Each node of the tree corresponds to the sequence of coordinates tested so far.
If the observed values do not determine $F$, the feasible actions are the tests of the untested coordinates.
Otherwise, the policy stops.
With $\Gamma=\sum_ic_i>0$, the reward at a leaf $\ell$ is $f_\ell=1-(\sum_{i\in\obs(\ell)}c_i)/\Gamma$.

To see that SBFE need not be FSD-monotone, consider the majority function of three variables with unit testing costs shown in Figure~\ref{fig:tree-example}, and suppose that all three variables have a common success probability $p\in(0,1)$.
At least two tests are required, and the third is needed exactly when the first two outcomes differ.
By symmetry, regardless of the order, the optimal expected cost is $2+2p(1-p)$.
For $p=1/4,1/2,3/4$, the Bernoulli distribution increases in FSD, while the cost is $19/8$, $5/2$, and $19/8$.
Thus, the problem is not FSD-monotone, although the majority function is monotone as a Boolean function.

\begin{proposition}\label{prop:sbfe}
For SBFE with any Boolean function and nonnegative testing costs, the expected regret in the original testing cost is $\otil(\Gamma\sqrt{nT})$.
\end{proposition}
\begin{proof}
Every coordinate is binary, and hence Assumption~\ref{ass:level-set} holds with $s_i=1$ for every $i$, as noted after the assumption.
Theorem~\ref{thm:local} with $S=n$ implies the claim.
\end{proof}
Let $C^*$ be the optimal expected testing cost.
Since the optimal normalized reward is $1-C^*/\Gamma$, Proposition~\ref{prop:first-order} bounds the regret in the original testing cost also by $\otil(\sqrt{nT\,C^*(\Gamma-C^*)}+\Gamma n)$, which is smaller when $C^*\ll\Gamma$.
For threshold functions $F(x)=\ind\{x_1+\cdots+x_n\ge\theta\}$, which include the majority function, the reward splits into two parts that are FSD-monotone in opposite directions, and a variant of the classical optimal policy for known distributions~\cite{BenDov1981,SalloumBreuer1984} is optimal for both parts under different optimistic distributions.
Applying Proposition~\ref{prop:single} to each part yields an algorithm with $O(n^2)$ time per round and regret $\otil(\Gamma\sqrt{nT})$ (Proposition~\ref{prop:sbfe-threshold} in Appendix~\ref{app:sbfe-threshold}).

\subsection{Full-Information Secretary Problem}\label{subsec:secretary}
In the full-information secretary problem, candidates with independent values $X_1,\ldots,X_n\in[0,1]$ arrive in this order, the decision maker decides immediately whether to accept the current candidate, and the goal is to maximize the probability of accepting a candidate with the largest value among all $n$ candidates.
Unlike in the classical secretary problem, the decision maker observes the values themselves rather than only their relative ranks.
Gilbert and Mosteller~\cite{gilbert1966recognizing} study identical known distributions, and Esfandiari, Hajiaghayi, Lucier, and Mitzenmacher~\cite{esfandiari2020prophets} study non-identical known distributions.
Harada, Kawase, and Sumita~\cite{harada2025repeated} study the repeated version with unknown distributions under full feedback, together with per-round guarantees relative to a baseline.
Since leaf rewards depend only on revealed values and the success of a policy depends on the values after the accepted candidate, these values must be revealed after acceptance, and hence all values are observed in every round.

Let $u_i$ be the node at time $i$, with $X_1$ initially observed at the root $r=u_1$.
For $i<n$, the node $u_i$ has a child $u_{i+1}$ revealing $X_{i+1}$ and another child that starts a chain of revelation edges revealing $X_{i+1},\ldots,X_n$ in order with no intermediate choices and ending at a leaf $\ell_i$, and $u_n=\ell_n$ is the final leaf.
The reward at $\ell_i$ is $f_{\ell_i}(x)=\ind\{x_i\ge x_j\text{ for all }j\in[n]\}$, and hence $\Gamma=1$.

The problem is not FSD-monotone even with two candidates.
Let $n=2$, $X_1=1/2$, and $X_2\in\{0,1\}$ with $\Prb(X_2=1)=q$.
Accepting the first candidate succeeds with probability $1-q$, and accepting the second succeeds with probability $q$.
Hence, the optimal success probability is $\max\{q,1-q\}$, which first decreases and then increases as $q$ increases.

\begin{proposition}\label{prop:secretary}
For the full-information secretary problem, Assumption~\ref{ass:level-set} holds with $s_1=\cdots=s_n=2$, and the expected regret is $\otil(\sqrt{nT})$.
\end{proposition}
\begin{proof}
Fix arbitrary nonempty sets of distributions as in Assumption~\ref{ass:level-set}.
Every leaf reward is nondecreasing in the value of the accepted candidate and nonincreasing in the values of the other candidates, and the feasible actions do not depend on the values.
Expectations, suprema, and maxima preserve these monotonicities.
Hence, backward induction shows that every optimistic continuation value on the chain after accepting candidate $i$ is nondecreasing in $x_i$ and nonincreasing in the values of the other candidates, and that the optimistic continuation value at $u_k$ is nonincreasing in each past value $x_j$ with $j<k$.
The optimistic continuation value after a value is revealed on a chain is therefore monotone in that value.
At $u_i$ with $i<n$, the value of accepting is nondecreasing in $x_i$, and the value of continuing is nonincreasing in $x_i$.
Their maximum has level-set complexity at most 2, and the value at $u_n$ is nondecreasing in $x_n$.
Thus, Assumption~\ref{ass:level-set} holds with $s_i=2$, and Theorem~\ref{thm:local} with $S\le2n$ implies the claim.
\end{proof}
Proposition~\ref{prop:first-order} refines the bound to $\otil(\sqrt{nT\,p^*(1-p^*)}+n)$, where $p^*$ is the optimal success probability, and this bound is smaller when $p^*$ is close to 0 or 1.
If every domain is finite, Proposition~\ref{prop:states} applies with the following states.
At $u_i$, the state is the pair of $\max_{j<i}x_j$, which is 0 for $i=1$, and $x_i$.
On the chain after accepting candidate $i$, the state is the pair of $x_i$ and the indicator that $x_i$ is at least every value revealed so far.
Each round then runs in time polynomial in $n$, $t$, and the sizes of the domains.

The model also captures random arrival orders.
For $k\in[n-1]$, let $Y_k$ be an additional coordinate with the uniform distribution on $[0,1]$.
Before the $k$th arrival, that is, at the root if $k=1$ and after the rejection of the $(k-1)$th arriving candidate otherwise, the next edge reveals $Y_k$.
The only feasible action then reveals the candidate selected by $Y_k$ through the conditional distribution of the $k$th arrival given the previous arrivals, for example, by listing the remaining candidates in a fixed order and selecting the first one whose cumulative conditional probability is at least $Y_k$.
When one candidate remains, rejecting reveals this candidate.
In this way, every known distribution over the $n!$ arrival orders arises, and for uniformly random orders, $Y_k$ can instead be uniform on $n-k+1$ values.
Since the distributions of the coordinates $Y_k$ are known, the algorithm can use the set consisting only of the distribution of $Y_k$ as its confidence set.
The gaps at the stages that reveal $Y_k$ then vanish in the proof of Theorem~\ref{thm:local}, and this proof, together with that of Proposition~\ref{prop:secretary}, which applies to every arrival order, yields regret $\otil(\sqrt{nT})$ with $S$ summing only over the candidates.
The decision tree has a node for each possible order of the candidates seen so far, but the regret bound does not depend on its size.

\subsection{Optimal Stopping with Loss Aversion}\label{subsec:behavioral}
Kleinberg, Kleinberg, and Oren~\cite{kleinberg2022optimal} study a variant of optimal stopping with loss aversion, in which the best value seen so far serves as a reference point and accepting a worse candidate incurs a penalty.
In our setting, independent candidate values $X_1,\ldots,X_n\in[0,1]$ with unknown distributions are observed in this order.
After observing each value, the policy either accepts the current candidate and stops or observes the next one, and the last candidate must be accepted.
Accepting candidate $i$ yields the utility $x_i-\lambda(\max_{j<i}x_j-x_i)_+$, where $\lambda\ge0$ is a known loss-aversion coefficient and the maximum over the empty set is 0.

Let $u_i$ be the node at time $i$, with $X_1$ initially observed at the root $r=u_1$.
For $i<n$, the node $u_i$ has a stopping leaf $\ell_i$ and a child $u_{i+1}$ revealing $X_{i+1}$, and $u_n=\ell_n$ is the final leaf.
Since utilities lie in $[-\lambda,1]$, we set $\Gamma=1+\lambda$ and define the leaf reward as
\begin{equation*}
f_{\ell_i}(x_1,\ldots,x_i)
=\frac{x_i-\lambda(\max_{j<i}x_j-x_i)_++\lambda}{1+\lambda}.
\end{equation*}

Both directions of FSD monotonicity can fail even with two candidates.
Let $\lambda>0$ and $n=2$.
First let $X_1=0$, $X_2\sim\operatorname{Bernoulli}(q)$, and $q\in(0,1)$.
The optimal expected utility is $q$.
If only the first distribution is made larger in FSD by setting $X_1=a$, $0<a<q$, then the optimal expected utility becomes $\max\{a,q-\lambda a(1-q)\}<q$.
Thus, making a distribution larger can decrease the optimal value.
On the other hand, with $X_1=0$, increasing a deterministic second candidate from $1/2$ to 1 increases the optimal value, and hence the problem is not FSD-monotone.

\begin{proposition}\label{prop:bias}
For optimal stopping with loss aversion and every $\lambda\ge0$, Assumption~\ref{ass:level-set} holds with $s_1=\cdots=s_n=2$, and the expected regret in the original utility is $\otil(\sqrt{(1+\lambda)nT}+(1+\lambda)n)$.
\end{proposition}
\begin{proof}
Fix arbitrary nonempty sets of distributions as in Assumption~\ref{ass:level-set}.
The reward for accepting candidate $i$ is nondecreasing in $x_i$ and nonincreasing in every past value, since increasing a past value cannot lower the reference point.
As in the proof of Proposition~\ref{prop:secretary}, the value of accepting the current candidate is nondecreasing in its value, the value of continuing is nonincreasing in it, and Assumption~\ref{ass:level-set} holds with $s_i=2$.

Let $U^*$ be the optimal expected utility, which lies in $[0,1]$ since accepting the first candidate yields nonnegative utility.
Proposition~\ref{prop:first-order} with $S\le2n$ and $p=(U^*+\lambda)/(1+\lambda)$ bounds the normalized regret by $\otil(\sqrt{nT\,p(1-p)}+n)$.
Multiplying by $\Gamma=1+\lambda$ and using $(1+\lambda)^2p(1-p)=(U^*+\lambda)(1-U^*)$ yields regret $\otil(\sqrt{nT\,(U^*+\lambda)(1-U^*)}+(1+\lambda)n)$ in the original utility, and $(U^*+\lambda)(1-U^*)\le1+\lambda$ proves the claim.
\end{proof}

The dependence on $\lambda$ is necessary already for two candidates.
\begin{proposition}\label{prop:bias-lower}
For every $\lambda\ge0$ and $T\ge1$, there are two instances of optimal stopping with loss aversion with $n=2$ candidates such that every learning algorithm incurs expected regret at least $\frac1{48}\min\{T,\sqrt{(\lambda+2)T}\}$ in the original utility on one of them, even under full feedback, in which both values are revealed at the end of every round.
\end{proposition}
For $n=2$ and $T\ge\lambda+2$, this lower bound is at least $\frac1{48}\sqrt{(1+\lambda)T}$, and hence the dependence of Proposition~\ref{prop:bias} on $\lambda$ and $T$ is tight up to logarithmic factors.
Appendix~\ref{app:proofs} proves Proposition~\ref{prop:bias-lower}.
Appendix~\ref{subsec:states} shows that each round runs in polynomial time when each domain is $[0,1]$ or finite.

\subsection{Stochastic Scheduling with Deadlines}\label{subsec:scheduling}
In stochastic scheduling with deadlines, a single machine processes $n$ jobs one at a time without preemption.
Job $i$ has a known release time $e_i\ge0$, a known deadline $d_i$, and a random outcome $X_i\in[0,1]$.
The outcomes are independent across jobs and have unknown distributions.
The processing time is $p_i(X_i)$, and completing the job by its deadline earns reward $r_i(X_i)$, where $p_i$ and $r_i$ are known nonnegative functions.
We assume that $p_i$ is nondecreasing and $r_i$ has level-set complexity at most an integer $s_i\ge1$.
We can take $s_i=1$ for fixed, monotone, or unimodal reward functions.
Examples include rewards proportional to processing time, $r_i(x)=a_i p_i(x)$, and capped rewards $r_i(x)=\min\{a_i p_i(x),b_i\}$ with $a_i,b_i\ge0$.
If job $i$ has finitely many possible processing-time--reward pairs, ordering the pairs by processing time and assigning distinct outcome labels allows us to take $s_i$ equal to half their number rounded up.

A policy starts at time 0 and repeatedly selects an unprocessed job or stops.
If job $i$ is selected when the machine becomes free at time $\tau$, it starts at time $\max\{\tau,e_i\}$ and completes at time $\max\{\tau,e_i\}+p_i(X_i)$.
The job runs to completion even if it misses its deadline, in which case it earns no reward.
Upon completion, the learner observes $(p_i(X_i),r_i(X_i))$ and may continue with another job.
We assume that the observed pair uniquely identifies $X_i$, including when the deadline is missed.

Let $\Gamma>0$ be a known upper bound on the total earned reward, for example, $\sum_i\sup_xr_i(x)$.
In the tree, each node records the sequence of jobs processed so far, and their observed processing times determine the current time.
The normalized leaf reward is the total earned reward divided by $\Gamma$.
Dean~\cite{dean2005thesis} studies stochastic scheduling with deadlines, including release times.
Here, processing time and reward may depend on a common random outcome.

Correlated stochastic knapsack~\cite{GuptaKMR11,Ma18}, under the conditions above, is a special case with $e_i=0$ and $d_i=B$ for every $i$, where $B>0$ is the capacity.
Each job corresponds to an item with size $p_i(X_i)$ and reward $r_i(X_i)$, earned if the cumulative size including that item is at most $B$.
As in the standard stochastic knapsack model~\cite{DeanGV08}, the process stops at the first overflow, and the overflowing item earns no reward.
Our analysis also covers the soft-constraint model~\cite{YoshinagaK25}, in which the overflowing item earns its reward before the process stops.
If $r_i(x)\le Lp_i(x)$ for all $i$ and $x$, we can take $\Gamma=LB$ under the standard convention and $\Gamma=LB+\max_i\sup_xr_i(x)$ under the soft-constraint convention.

To see that FSD monotonicity can fail even under the standard knapsack convention, consider capacity $B=1$ and two items.
For item 1, take $p_1(z)=r_1(z)=z$ and let $X_1=x$ deterministically, while item 2 has fixed size $1/2$ and reward 1.
For $x=1/4,1/2,3/4$, the known-distribution optimal total rewards are, respectively, $5/4$, $3/2$, and $1$.
In the first two cases both items fit, while in the last case selecting only item 2 is optimal.
The distribution of $X_1$ increases in FSD, but the optimal value first increases and then decreases.
Thus, the problem is not FSD-monotone.

We apply Theorem~\ref{thm:sum} by separating the reward of the current job from the value of the remaining jobs.
The latter is nonincreasing in the current outcome because $p_i$ is nondecreasing.
Their sum need not have small level-set complexity, even when the immediate reward is monotone.
For example, consider knapsack with $B=1$, item 1 with $p_1(z)=r_1(z)=z$, and item 2 with reward 1 and size uniform on $\{1/m,2/m,\ldots,1\}$.
After item 1 is inserted, the optimal continuation value is proportional to $x_1+\lfloor m(1-x_1)\rfloor/m$, which equals 1 at $x_1\in\{0,1/m,\ldots,1\}$ and lies in $(1-1/m,1)$ elsewhere, and hence its superlevel set at level $1-1/(2m)$ consists of $m+1$ separated parts.

\begin{proposition}\label{prop:scheduling}
For the scheduling model above with nondecreasing processing-time functions $p_i$ and reward functions $r_i$ of level-set complexity at most $s_i$, let $S=\sum_i s_i$.
The expected regret in the original total reward is $\otil(\Gamma\sqrt{ST})$.
\end{proposition}
\begin{proof}
Fix an edge revealing job $i$, its start time $\tau_i$, and all preceding observations.
For each leaf below this edge, write $f_\ell=\kappa+\chi(x_i)+F_\ell$, where $\kappa$ and $F_\ell$ are the normalized rewards earned before and after job $i$, and $\chi(z)=\ind\{\tau_i+p_i(z)\le d_i\}r_i(z)/\Gamma$.
The reward bound ensures that $\chi,F_\ell\in[0,1]$.
Since $p_i$ is nondecreasing, the set $\{z:\tau_i+p_i(z)\le d_i\}$ is an initial interval of the outcome domain.
At a nonnegative level, a superlevel set of $\chi$ is the intersection of this set with a superlevel set of $r_i$, and hence has at most $s_i$ interval components.

Increasing $x_i$ cannot decrease $p_i(x_i)$.
Thus, no subsequent job along a fixed path completes earlier, and $F_\ell$ cannot increase.
The sets of feasible actions remain unchanged, except that an overflow in the knapsack case allows only stopping.
The immediate rewards sum to the normalized leaf reward and hence to at most 1.
Theorem~\ref{thm:sum} and rescaling by $\Gamma$ imply the claim.
For the soft-constraint knapsack model, the same proof applies with $\chi(z)=r_i(z)/\Gamma$.
\end{proof}
Let $\opt^*$ be the optimal expected total reward.
Since the immediate rewards sum to the leaf reward, Proposition~\ref{prop:sum-first-order} with $p=\opt^*/\Gamma$ improves the bound to $\otil(\sqrt{ST\,\Gamma\opt^*}+\Gamma S)$.
When the $s_i$ are bounded by a constant, both bounds hold with $n$ in place of $S$, up to constant factors.

\subsection{Overbooking with Group Reservations}\label{subsec:overbooking}
In overbooking with group reservations, a service provider receives reservation requests from $n$ groups.
Group $i$ has $b_i\ge1$ members, pays a fee $a_i\ge0$, and receives a refund $\rho_i\ge0$ for each member who does not show up, where $\rho_ib_i\le a_i$.
The number of members of group $i$ who show up is $Y_i\in\{0,\ldots,b_i\}$.
These counts are independent across groups and have unknown distributions.
We make no assumption on the dependence between the members of a group.
The capacity is an integer $B\ge0$, and each customer beyond the capacity incurs a cost $c\ge0$.
The parameters $b_i$, $a_i$, $\rho_i$, $B$, and $c$ are known.
We take $X_i=Y_i/b_i$ as the coordinate of group $i$, whose domain is $\{0,1/b_i,2/b_i,\ldots,1\}$.
The case $b_1=\cdots=b_n=1$ is the model of individual reservations with binary show-up outcomes.
Reservation management with cancellations, refunds, and no-shows is studied, for example, by Subramanian, Stidham, and Lautenbacher~\cite{subramanian1999}.

Before any show-up outcome is observed, the policy chooses a set $U$ of accepted reservations from a known family $\mathcal U\subseteq2^{[n]}$ with $\emptyset\in\mathcal U$.
At the time of service, it observes $Y_i$ for $i\in U$ and receives the net revenue
\begin{align*}
W(U,Y_U)=\sum_{i\in U}\bigl(a_i-\rho_i(b_i-Y_i)\bigr)
-c\Bigl(\sum_{i\in U}Y_i-B\Bigr)_+.
\end{align*}
Since the policy chooses $U$ before observing any outcome, this is a combinatorial semi-bandit problem.
The monotonicity condition for combinatorial bandits with general reward functions~\cite{chen2016general} can fail here because of the cost of exceeding capacity.

In the tree, choosing a nonempty set $U$ reveals its coordinates along a chain in a fixed order, while choosing $U=\emptyset$ leads directly to a stopping leaf.
Let $\bar b=\sum_i b_i$ and $C=c(\bar b-B)_+$.
Since $-C\le W(U,Y_U)\le\sum_i a_i$, we take $\Gamma=\sum_i a_i+C$, assume $\Gamma>0$, and normalize the leaf reward as $f_\ell=(W(U,y_U)+C)/\Gamma$.

The optimal value can also be non-monotone in the show-up distribution.
For example, consider a single group with $b_1=3$, $a_1=3$, and $\rho_1=1$, and let $B=2$ and $c=2$.
If $Y_1$ is deterministic and equal to $1$, $2$, or $3$, accepting the group yields the net revenue $1$, $2$, or $1$, respectively, while rejecting it yields 0.
Thus, the optimal value first increases and then decreases, and the problem is not FSD-monotone.

We apply Theorem~\ref{thm:sum} by decomposing the net revenue into the refund savings for the current group and the remaining net revenue.
The former is nondecreasing in the current show-up count, while the latter is nonincreasing.
Their sum, the full optimistic continuation value, can have level-set complexity that grows with the group sizes.
Let $b_1=b_2=B=a_1=a_2=2m$, $\rho_1=\rho_2=1$, and $c=2$, accept both groups, and let the set of distributions of $Y_2$ consist of the point masses at $0,2,\ldots,2m$.
Then, the optimistic net revenue after $Y_1=y$ is observed is the maximum of $y+y'-2(y+y'-2m)_+$ over these values $y'$, which equals $2m$ for even $y$ and $2m-1$ for odd $y$.
Hence, its superlevel set at level $2m-1/2$ consists of $m+1$ separated points, and Assumption~\ref{ass:level-set} would require $s_1\ge m+1$, which grows with the group sizes.

\begin{proposition}\label{prop:overbooking}
For overbooking with group reservations, the expected regret in the original net revenue is $\otil(\Gamma\sqrt{nT})$, where $n$ is the number of groups.
\end{proposition}
Appendix~\ref{app:proofs} proves Proposition~\ref{prop:overbooking}.

Let $W^*$ be the optimal expected net revenue.
The total immediate reward is at most the normalized leaf reward because $a_i-\rho_i b_i\ge0$ and the overflow cost is at most $C$.
Proposition~\ref{prop:sum-first-order} with $p=(W^*+C)/\Gamma$ therefore improves the bound to $\otil(\sqrt{nT\,\Gamma(W^*+C)}+\Gamma n)$.
Apart from $\Gamma$, the basic bound does not depend on the group sizes $b_i$.
Appendix~\ref{subsec:states} shows that each round runs in time polynomial in $n$, $t$, $|\mathcal U|$, and $\bar b$.

\section{Analysis of the Upper Bound}\label{sec:analysis}
In this section, we prove the regret upper bounds of Section~\ref{sec:algorithm}.
Section~\ref{subsec:coverage} establishes confidence-set coverage and bounds the resulting expectation errors.
Section~\ref{subsec:main-upper} combines these bounds with a variance budget to prove Theorem~\ref{thm:local} and Propositions~\ref{prop:single} and~\ref{prop:first-order}.
Section~\ref{subsec:sums} extends the analysis to immediate rewards.

\subsection{Properties of Confidence Sets}\label{subsec:coverage}
To analyze adaptively collected samples, we represent the observations of each coordinate by an independent sequence of i.i.d.\ samples.
For each coordinate $i$, fix an infinite sequence $Z_{i,1},Z_{i,2},\ldots$ of i.i.d.\ random variables with distribution $D_i$, independently across coordinates, and interpret the $j$-th revealed value of coordinate $i$ as $Z_{i,j}$.
This preserves the observation law because coordinates are independent and each revelation decision precedes observing that coordinate's value.

The following interval bound is a consequence of the empirical Bernstein inequality of Maurer and Pontil~\cite[Theorem~6]{mp2009}, and Appendix~\ref{app:proofs} proves it.
\begin{lemma}\label{lem:interval-bernstein}
Let $m\ge16$ and $\eta\in(0,1)$, and let $\widehat P_m$ be the empirical distribution of i.i.d.\ samples $Z_1,\ldots,Z_m$ from a distribution $P$ on $[0,1]$.
With probability at least $1-\eta$, we have simultaneously for all $I\in\II_1$
\begin{equation}\label{eq:mp}
|P(I)-\widehat P_m(I)|
\le\sqrt{\frac{18\widehat P_m(I)(1-\widehat P_m(I))\,\beta}{m-1}}
+\frac{15\,\beta}{m-1},
\qquad\text{where }\beta=\log\frac{20(2m+1)^2}{\eta}.
\end{equation}
\end{lemma}

We now apply this bound to show that the confidence sets contain the true distributions with high probability.
\begin{lemma}\label{lem:coverage}
With probability at least $1-\delta$, we have $D_i\in\CC_m(\widehat D_{i,m})$ simultaneously for every $i\in[n]$ and every $0\le m\le T$.
\end{lemma}
\begin{proof}
For $m=0$, the inclusion holds by definition.
For $1\le m<16$, we have $b_m(z)=1$ for every $z\in[0,1]$ since $\Lambda>1$, and hence the constraints are vacuous.

Fix $i$ and $16\le m\le T$, and apply~\eqref{eq:mp} to $D_i$ and its samples $Z_{i,1},\ldots,Z_{i,m}$ with $\eta=\frac{\delta}{n(T+1)}$.
Since $(2m+1)^2\le4(T+1)^2$, we have $\beta=\log(20n(T+1)(2m+1)^2/\delta)\le\log(80n(T+1)^3/\delta)=\Lambda$.

For $I\in\II_1$, define $z=\widehat D_{i,m}(I)$.
By~\eqref{eq:mp}, we have
\begin{align*}
|D_i(I)-\widehat D_{i,m}(I)|
&\le\sqrt{\frac{18z(1-z)\beta}{m-1}}+\frac{15\beta}{m-1}
\le\sqrt{\frac{18z(1-z)\Lambda}{m-1}}+\frac{15\Lambda}{m-1}\\
&=\sqrt{\frac{18m}{m-1}}\sqrt{z(1-z)\frac{\Lambda}{m}}
+\frac{15m}{m-1}\frac{\Lambda}{m}
\le6\sqrt{z(1-z)\frac{\Lambda}{m}}+32\frac{\Lambda}{m}.
\end{align*}
The last inequality uses $m/(m-1)\le2$, which implies $\sqrt{18m/(m-1)}\le6$ and $15m/(m-1)\le30\le32$.
Since the difference is also at most 1, $D_i$ satisfies the constraints with the radius~\eqref{eq:radius}.
For each pair $(i,m)$ with $i\in[n]$ and $16\le m\le T$, the inclusion fails with probability at most $\delta/[n(T+1)]$.
A union bound over at most $n(T+1)$ such pairs, together with the deterministic inclusion for $m<16$, proves that all inclusions hold simultaneously with probability at least $1-\delta$.
\end{proof}

The next lemma bounds the integral that arises when a function is written using its superlevel sets.
\begin{lemma}\label{lem:layer}
Let $Z\in[0,1]$ be a random variable, and define $p_y=\Prb(Z>y)$.
For every $\eta\in(0,1)$,
\begin{equation}\label{eq:layer-variance}
\int_0^1\sqrt{p_y(1-p_y)}\,dy
\le\eta+\sqrt{\Var(Z)\log(1/\eta)}.
\end{equation}
\end{lemma}
\begin{proof}
Let $\mu=\E[Z]$.
We split the integral in~\eqref{eq:layer-variance} into the regions $|y-\mu|<\eta$ and $|y-\mu|\ge\eta$.
The first part is at most $\eta$, since its integration range has length at most $2\eta$ and $\sqrt{p_y(1-p_y)}\le1/2$ for all $p_y\in[0,1]$.
For the second part, the Cauchy--Schwarz inequality implies
\[
\int_{\{y\in[0,1]:|y-\mu|\ge\eta\}}\sqrt{p_y(1-p_y)}\,dy
\le\left(\int_0^1|y-\mu|\,p_y(1-p_y)\,dy\right)^{1/2}
\left(\int_{\{y\in[0,1]:|y-\mu|\ge\eta\}}\frac{dy}{|y-\mu|}\right)^{1/2}.
\]

For the first integral, we use $p_y(1-p_y)\le p_y=\Prb(Z>y)$ for $y\ge\mu$ and $p_y(1-p_y)\le(1-p_y)=\Prb(Z\le y)$ for $y<\mu$.
Then, we have
\begin{align*}
\int_0^1|y-\mu|\,p_y(1-p_y)\,dy
&\le\int_\mu^1(y-\mu)\Prb(Z>y)\,dy+\int_0^\mu(\mu-y)\Prb(Z\le y)\,dy\\
&=\frac{\E[(Z-\mu)_+^2]+\E[(\mu-Z)_+^2]}{2}
=\frac{\E[(Z-\mu)^2]}2
=\frac{\Var(Z)}2.
\end{align*}
The first equality follows from the tail-integral formula $\E[U^2]=2\int_0^\infty t\Prb(U>t)\,dt$ for nonnegative $U$, applied to $(Z-\mu)_+$ and $(\mu-Z)_+$.

For the second integral, substitute $u=|y-\mu|$ separately on each side of $\mu$.
The integration range on either side is contained in $[\eta,1]$, and hence
\[
\int_{\{y\in[0,1]:|y-\mu|\ge\eta\}}\frac{dy}{|y-\mu|}
\le2\int_\eta^1\frac{du}{u}
=2\log(1/\eta).
\]
The product of the two square roots is therefore at most $\sqrt{\Var(Z)\log(1/\eta)}$.
Adding the contribution of at most $\eta$ from $|y-\mu|<\eta$ proves the claim.
\end{proof}

The following lemma compares two distributions in the same confidence set, such as $D_i$ and a distribution used in the optimistic update, first on unions of intervals and then on functions with small level-set complexity.
\begin{lemma}\label{lem:level-stability}
Let $s\ge1$ and $m\ge0$ be integers and $P,Q\in\CC_m(\widehat D_{i,m})$.
\begin{enumerate}[label=(\alph*),nosep]
\item If $m\ge1$, then for every $B\in\II_s$,
\begin{equation}\label{eq:two-distributions-union}
|Q(B)-P(B)|
\le24\sqrt{P(B)(1-P(B))\frac{s\Lambda}{m}}
+368\frac{s\Lambda}{m}.
\end{equation}
\item For every measurable $[0,1]$-valued function $\phi$ with level-set complexity at most $s$,
\begin{equation*}
|\E_{X\sim Q}[\phi(X)]-\E_{X\sim P}[\phi(X)]|
\le24\sqrt{\frac{s\Lambda\log(m+1)}{m+1}\Var_{X\sim P}(\phi(X))}
+\frac{417s\Lambda}{m+1}.
\end{equation*}
\end{enumerate}
\end{lemma}
\begin{proof}
For $m=0$, part~(b) holds because its last term exceeds 1.
For $1\le m<16$, both bounds hold because their left-hand sides are at most 1 and their last terms exceed 1.

Let $m\ge16$.
For~(a), we first compare the empirical variance with the variance under $P$ on an interval $I\in\II_1$.
Define $z=\widehat D_{i,m}(I)$, $p=P(I)$, and $w=\sqrt{z(1-z)}$.
Since $z,p\in[0,1]$, we have
\[
z(1-z)-p(1-p)
\le|z(1-z)-p(1-p)|
=|z-p|\,|1-z-p|
\le|z-p|.
\]
Since $P$ belongs to the confidence set, we obtain
\[
w^2=z(1-z)
\le p(1-p)+|z-p|
\le p(1-p)+6\sqrt{z(1-z)\frac{\Lambda}{m}}+32\frac{\Lambda}{m}
=p(1-p)+6w\sqrt{\frac{\Lambda}{m}}+32\frac{\Lambda}{m}.
\]
Completing the square yields $(w-3\sqrt{\Lambda/m})^2\le p(1-p)+41\Lambda/m$, and hence
\begin{equation}\label{eq:empirical-standard-deviation}
w\le3\sqrt{\frac{\Lambda}{m}}+\sqrt{p(1-p)+\frac{41\Lambda}{m}}
\le\sqrt{p(1-p)}+(3+\sqrt{41})\sqrt{\frac{\Lambda}{m}}
\le\sqrt{p(1-p)}+10\sqrt{\frac{\Lambda}{m}},
\end{equation}
where the last two inequalities use $\sqrt{x+y}\le\sqrt{x}+\sqrt{y}$ for $x,y\ge0$ and $3+\sqrt{41}\le10$, respectively.
By the definition of our confidence set for $P$ and $Q$ and~\eqref{eq:empirical-standard-deviation}, we now have
\begin{equation}\label{eq:two-distributions}
\begin{aligned}
|Q(I)-P(I)|
&\le|Q(I)-\widehat D_{i,m}(I)|+|P(I)-\widehat D_{i,m}(I)|\\
&\le12\sqrt{z(1-z)\frac{\Lambda}{m}}+64\frac{\Lambda}{m}
\le12\sqrt{p(1-p)\frac{\Lambda}{m}}+184\frac{\Lambda}{m}.
\end{aligned}
\end{equation}

We next extend~\eqref{eq:two-distributions} to a union $B\in\II_s$.
Write $B$ as a disjoint union of intervals $I_1,\ldots,I_k$ with $k\le s$, and define $p_r=P(I_r)$ and $p=P(B)=\sum_{r=1}^k p_r$.
Applying~\eqref{eq:two-distributions} to each interval and using $p_r(1-p_r)\le p_r$, we obtain
\begin{equation}\label{eq:union-mass-bound}
|Q(B)-P(B)|
\le\sum_{r=1}^k|Q(I_r)-P(I_r)|
\le12\sqrt{\frac{\Lambda}{m}}\sum_{r=1}^k\sqrt{p_r}+184\frac{k\Lambda}{m}
\le12\sqrt{\frac{sp\Lambda}{m}}+184\frac{s\Lambda}{m}.
\end{equation}
The last inequality follows from the arithmetic-quadratic mean inequality, since $\sum_{r=1}^k\sqrt{p_r}\le\sqrt{k\sum_{r=1}^k p_r}=\sqrt{kp}\le\sqrt{sp}$.
The complement $B^c=[0,1]\setminus B$ is a disjoint union of at most $s+1$ intervals and satisfies $P(B^c)=1-p$.
Applying the same argument to $B^c$ and using $|Q(B^c)-P(B^c)|=|Q(B)-P(B)|$ yields
\begin{equation}\label{eq:complement-mass-bound}
|Q(B)-P(B)|\le12\sqrt{\frac{(s+1)(1-p)\Lambda}{m}}+184\frac{(s+1)\Lambda}{m}.
\end{equation}
Since $p\in[0,1]$, we have $\min\{p,1-p\}\le2p(1-p)$.
Together with $s+1\le2s$, this shows that the smaller of the bounds in~\eqref{eq:union-mass-bound} and~\eqref{eq:complement-mass-bound} is at most the right-hand side of~\eqref{eq:two-distributions-union}.

For~(b), define $B_y=\{z\in\operatorname{dom}(\phi):\phi(z)>y\}$ and $p_y=P(B_y)$.
By assumption, each $B_y$ is the intersection of $\operatorname{dom}(\phi)$ with a set in $\II_s$.
Using $\phi(x)=\int_0^1\ind\{\phi(x)>y\}\,dy$ and applying~\eqref{eq:two-distributions-union} to the set in $\II_s$ whose intersection with $\operatorname{dom}(\phi)$ is $B_y$, which has the same probabilities under $P$ and $Q$ since both are supported on $\operatorname{dom}(\phi)$, we obtain
\begin{align*}
|\E_{X\sim Q}[\phi(X)]-\E_{X\sim P}[\phi(X)]|
&=\left|\int_0^1\bigl(Q(B_y)-P(B_y)\bigr)\,dy\right|
\le\int_0^1|Q(B_y)-P(B_y)|\,dy\\
&\le24\sqrt{\frac{s\Lambda}{m}}\int_0^1\sqrt{p_y(1-p_y)}\,dy
+\frac{368s\Lambda}{m}.
\end{align*}
Applying Lemma~\ref{lem:layer} to $Z=\phi(X)$ for $X\sim P$ with $\eta=(m+1)^{-1/2}$, we obtain
\[
\int_0^1\sqrt{p_y(1-p_y)}\,dy
\le\frac{1}{\sqrt{m+1}}+\sqrt{\frac{\Var_{X\sim P}(\phi(X))\log(m+1)}{2}}.
\]
Substituting this estimate into the preceding bound yields
\begin{align*}
|\E_{X\sim Q}[\phi(X)]-\E_{X\sim P}[\phi(X)]|
&\le24\sqrt{\frac{s\Lambda\log(m+1)}{2m}\Var_{X\sim P}(\phi(X))}
+\frac{24\sqrt{s\Lambda}}{\sqrt{m(m+1)}}+\frac{368s\Lambda}{m}\\
&\le24\sqrt{\frac{s\Lambda\log(m+1)}{m+1}\Var_{X\sim P}(\phi(X))}
+\frac{392s\Lambda}{m}\\
&\le24\sqrt{\frac{s\Lambda\log(m+1)}{m+1}\Var_{X\sim P}(\phi(X))}
+\frac{417s\Lambda}{m+1}.
\end{align*}
The second inequality uses $2m\ge m+1$, $\sqrt{m(m+1)}\ge m$, and $\sqrt{s\Lambda}\le s\Lambda$.
The last inequality follows from $(m+1)/m\le17/16$ and $392\cdot17/16=416.5\le417$.
\end{proof}

When $P$ has few support points, the following one-sided version of Lemma~\ref{lem:level-stability}(b) holds for every function, which suffices for the optimistic updates.
\begin{lemma}\label{lem:finite-support}
Let $k\ge1$ and $m\ge0$ be integers, and let $P,Q\in\CC_m(\widehat D_{i,m})$, where $P$ has at most $k$ support points.
Then, for every measurable $[0,1]$-valued function $\phi$,
\[
\E_{X\sim Q}[\phi(X)]-\E_{X\sim P}[\phi(X)]
\le24\sqrt{\frac{k\Lambda\log(m+1)}{m+1}\Var_{X\sim P}(\phi(X))}
+\frac{417k\Lambda}{m+1}.
\]
\end{lemma}
\begin{proof}
Let $A$ be the support of $P$, and let $\psi(z)=(1-\phi(z))\ind\{z\in A\}$.
Since $\psi\ge0$, every superlevel set of $\psi$ at a negative level is the whole domain, and every superlevel set at a level $y\ge0$ is a subset of $A$ and hence the intersection of the domain with a union of at most $k$ singletons.
Thus, $\psi$ has level-set complexity at most $k$.
Since $\phi\le1-\psi$, with equality on $A$, and $P(A)=1$, we have
\[
\E_{X\sim Q}[\phi(X)]-\E_{X\sim P}[\phi(X)]
\le\E_{X\sim P}[\psi(X)]-\E_{X\sim Q}[\psi(X)]
\]
and $\Var_{X\sim P}(\psi(X))=\Var_{X\sim P}(1-\phi(X))=\Var_{X\sim P}(\phi(X))$.
Lemma~\ref{lem:level-stability}(b) applied to $\psi$ with $s=k$ proves the claim.
\end{proof}

\subsection{Proof of the Main Upper Bound}\label{subsec:main-upper}
Before proving Theorem~\ref{thm:local}, we bound the expected sum of the conditional variances of a process with values in $[0,1]$ in terms of the positive parts of its drifts.
\begin{lemma}\label{lem:variance-budget}
Let $(Y_h)_{h=0}^H$ be a process adapted to a filtration $(\FF_h)_{h=0}^H$ and taking values in $[0,1]$, and let $d_h=Y_h-\E[Y_{h+1}\mid\FF_h]$ for $0\le h<H$.
Then, we have
\[
\E\biggl[\sum_{h=0}^{H-1}\Var(Y_{h+1}\mid\FF_h)\biggr]
\le\E[Y_H^2]-\E[Y_0^2]+2\E\biggl[\sum_{h=0}^{H-1}(d_h)_+\biggr].
\]
If, in addition, $(Y_h)$ is a supermartingale, that is, $d_h\ge0$ for every $h$, then the right-hand side is at most 1.
\end{lemma}
\begin{proof}
Since $\E[Y_{h+1}\mid\FF_h]=Y_h-d_h$, we have
\begin{align*}
\Var(Y_{h+1}\mid\FF_h)
&=\E[Y_{h+1}^2\mid\FF_h]-(Y_h-d_h)^2
=\E[Y_{h+1}^2\mid\FF_h]-Y_h^2+d_h(2Y_h-d_h)\\
&\le\E[Y_{h+1}^2\mid\FF_h]-Y_h^2+2(d_h)_+,
\end{align*}
where the inequality follows from $d_h(2Y_h-d_h)\le2d_h$ if $d_h\ge0$, since $Y_h\le1$, and from $d_h(2Y_h-d_h)\le0$ if $d_h<0$, since $Y_h\ge0$.
By the tower property, $\E[\E[Y_{h+1}^2\mid\FF_h]]=\E[Y_{h+1}^2]$.
Taking expectations and summing over $0\le h<H$, we obtain the first claim, since the first sum telescopes:
\[
\E\biggl[\sum_{h=0}^{H-1}\Var(Y_{h+1}\mid\FF_h)\biggr]
\le\sum_{h=0}^{H-1}\bigl(\E[Y_{h+1}^2]-\E[Y_h^2]\bigr)+2\E\biggl[\sum_{h=0}^{H-1}(d_h)_+\biggr]
=\E[Y_H^2]-\E[Y_0^2]+2\E\biggl[\sum_{h=0}^{H-1}(d_h)_+\biggr].
\]
If $d_h\ge0$ for every $h$, the tower property yields $\E[(d_h)_+]=\E[d_h]=\E[Y_h]-\E[Y_{h+1}]$, and hence the last sum is $\E[Y_0-Y_H]$.
Let $\mu=\E[Y_0]$.
Since $Y_H\in[0,1]$, we have $Y_H^2\le Y_H$, and $\E[Y_0^2]\ge\mu^2$ by Jensen's inequality.
Hence,
\[
\E[Y_H^2]-\E[Y_0^2]+2\E[Y_0-Y_H]
\le\E[Y_H]-\mu^2+2\mu-2\E[Y_H]
=1-(1-\mu)^2-\E[Y_H]
\le1.\qedhere
\]
\end{proof}

We are now ready to prove Theorem~\ref{thm:local}.
\begin{proof}[Proof of Theorem~\ref{thm:local}]
Let $G_t$ be the event that $D_i\in\CC_{i,t}$ for every $i\in[n]$.
This event is determined by the samples collected before round $t$.
We call round $t$ good if $G_t$ occurs.
By Lemma~\ref{lem:coverage}, we have $\Prb(G_t^c)\le\delta$ for every $t$.
Since $0\le\reg_t\le1$, we have
\begin{equation}\label{eq:regret-good-bad}
R_T=\sum_{t=1}^T\E[\reg_t]\le\sum_{t=1}^T\E[\ind_{G_t}\reg_t]+\delta T.
\end{equation}

We first bound the expected regret in a single good round.
Fix the history before round $t$ for which $G_t$ holds.
In this part of the proof, expectations are conditional on this history.
Since the value vector $X^t=(X_1^t,\ldots,X_n^t)\sim D$ drawn in round $t$ is independent of the history, these conditional expectations are taken over $X^t\sim D$.
At a leaf, the optimistic and true values are equal.
Since $D_i\in\CC_{i,t}$ at every revelation edge, backward induction shows that
\begin{equation}\label{eq:optimism}
\bar g_u^{\,t}(x_{\obs(u)})\ge g_u(x_{\obs(u)})
\end{equation}
at every node.
The supremum over $Q$ is at least the expectation under $D_i$, and taking the maximum over the same feasible actions preserves the inequality.

Execute the policy $\sigma^t$ selected by Algorithm~\ref{alg:main} in round $t$ under $D$, and let $u_{t,0},u_{t,1},\ldots,u_{t,h^*}$ be the visited nodes, where $u_{t,0}=r$ and $u_{t,h^*}=\ell^t$.
Since there are at most $n$ revelation edges and at most one stopping edge on the path, we have $h^*\le n+1$.
For $0\le h\le h^*$, let $\FF_{t,h}$ be the $\sigma$-algebra generated by the visited nodes and observed values up to arrival at $u_{t,h}$.
For $h^*<h\le n+1$, set $u_{t,h}=\ell^t$ and $\FF_{t,h}=\FF_{t,h^*}$.
Then, $\FF_{t,0}\subseteq\cdots\subseteq\FF_{t,n+1}$ is a filtration.
Define
\[
Y_{t,h}=\bar g_{u_{t,h}}^{\,t}(X_{\obs(u_{t,h})}^t),
\qquad 0\le h\le n+1.
\]
In particular, $Y_{t,0}=\bar g_r^{\,t}(X_{\obs(r)}^t)$ and $Y_{t,h}=f(\sigma^t,X^t)$ for $h^*\le h\le n+1$.
For $0\le h\le n$, define
\[
d_{t,h}=Y_{t,h}-\E[Y_{t,h+1}\mid\FF_{t,h}],
\qquad w_{t,h}=\Var(Y_{t,h+1}\mid\FF_{t,h}).
\]

Suppose $h<h^*$ and the chosen edge $(u_{t,h},u_{t,h+1})=(u,v)$ reveals coordinate $i$, with preceding observations $x_{\obs(u)}$.
The edge is chosen by~\eqref{eq:policy-choice}, while the unobserved coordinate still follows $D_i$ conditional on $\FF_{t,h}$.
Therefore, since $D_i\in\CC_{i,t}$,
\begin{align*}
d_{t,h}
&=\sup_{Q\in\CC_{i,t}}\E_{X_i\sim Q}[\bar g_v^{\,t}(X_i,x_{\obs(u)})]
-\E_{X_i\sim D_i}[\bar g_v^{\,t}(X_i,x_{\obs(u)})]\ge0,\\
w_{t,h}
&=\Var_{X_i\sim D_i}(\bar g_v^{\,t}(X_i,x_{\obs(u)})).
\end{align*}
If $h<h^*$ and the chosen edge is a stopping edge, its reward equals $Y_{t,h}$, and hence $Y_{t,h+1}=Y_{t,h}$.
This equality also holds for $h^*\le h\le n$ by definition.
Thus, $d_{t,h}=w_{t,h}=0$ in both cases.
Since the optimistic DP preserves the range $[0,1]$, every $Y_{t,h}$ lies in $[0,1]$.
Since $d_{t,h}\ge0$ for every $h$, the sequence $Y_{t,0},\ldots,Y_{t,n+1}$ is a supermartingale with respect to $(\FF_{t,h})_h$, and Lemma~\ref{lem:variance-budget} yields
\begin{equation}\label{eq:local-round-budget}
\E\biggl[\sum_{h=0}^{n}w_{t,h}\biggr]\le1.
\end{equation}

By the definition of $d_{t,h}$ and the tower property, we have $\E[\sum_{h=0}^{n}d_{t,h}]=\E[Y_{t,0}]-f(\sigma^t\mid D)$.
By~\eqref{eq:optimism} and~\eqref{eq:opt}, $\E[Y_{t,0}]\ge\opt(D)$, and hence
\begin{equation}\label{eq:local-round-regret}
\reg_t\le\E\biggl[\sum_{h=0}^{n}d_{t,h}\biggr].
\end{equation}

Let $\bar s_i=\min\{s_i,k_i\}$, and consider a stage $h<h^*$ at which the chosen edge $(u,v)$ reveals coordinate $i$, where $D_i\in\CC_{i,t}$ since $G_t$ holds.
If $s_i\le k_i$, then $z\mapsto\bar g_v^{\,t}(z,x_{\obs(u)})$ has level-set complexity at most $s_i$ by Assumption~\ref{ass:level-set}, and we apply Lemma~\ref{lem:level-stability}(b) to this function.
Otherwise, we apply Lemma~\ref{lem:finite-support} with $k=k_i$.
In both cases, we take $m=N_{i,t}$, $P=D_i$, and every $Q\in\CC_{i,t}$, and using $\log(N_{i,t}+1)\le\log T$, we obtain
\begin{equation}\label{eq:local-error}
0\le d_{t,h}\le24\sqrt{\frac{\bar s_i\Lambda\log T}{N_{i,t}+1}w_{t,h}}+\frac{417\bar s_i\Lambda}{N_{i,t}+1}.
\end{equation}

We then bound the cumulative expected regret.
From now on, expectations also average over the histories before the rounds.
Let $I_{i,t}$ indicate whether coordinate $i$ is revealed on an edge in round $t$.
If $I_{i,t}=1$ and coordinate $i$ is revealed at stage $h$, define $\widetilde w_{i,t}=w_{t,h}$.
If $I_{i,t}=0$, set $\widetilde w_{i,t}=0$.
Since each coordinate is revealed at most once per round and $w_{t,h}=0$ at stages without a revelation, we have $\sum_{i=1}^n\widetilde w_{i,t}I_{i,t}=\sum_{h=0}^{n}w_{t,h}$.
For a coordinate $i\notin\obs(r)$, every revelation occurs on an edge, and $N_{i,t+1}=N_{i,t}+I_{i,t}$ for every realization.
Hence, at the $k$-th round with $I_{i,t}=1$, we have $N_{i,t}+1=k$.
For $i\in\obs(r)$, we have $I_{i,t}=0$.
Thus, for every coordinate $i$ and every realization,
\begin{equation}\label{eq:harmonic-log}
\sum_{t=1}^T\frac{I_{i,t}}{N_{i,t}+1}\le\sum_{k=1}^T\frac1k\le1+\log T.
\end{equation}

Let $\Delta$ and $W$ be the cumulative gaps and conditional variances in good rounds,
\[
\Delta=\sum_{t=1}^T\E\biggl[\ind_{G_t}\sum_{h=0}^nd_{t,h}\biggr],
\qquad
W=\sum_{t=1}^T\E\biggl[\ind_{G_t}\sum_{h=0}^nw_{t,h}\biggr].
\]
By~\eqref{eq:local-round-regret}, we have $\sum_t\E[\ind_{G_t}\reg_t]\le\Delta$.
Since~\eqref{eq:local-round-budget} holds for every history before round $t$ for which $G_t$ occurs, we have
\begin{equation}\label{eq:local-budget-averaged}
W=\sum_{t=1}^T\E\biggl[\ind_{G_t}\sum_{h=0}^nw_{t,h}\biggr]\le\sum_{t=1}^T\Prb(G_t)\le T.
\end{equation}

We next bound the cumulative contribution of the square-root term in~\eqref{eq:local-error}.
We have
\begin{equation}\label{eq:cs-first}
\begin{aligned}
\MoveEqLeft\E\biggl[\sum_{t,i}\ind_{G_t}I_{i,t}\sqrt{\frac{\bar s_i\widetilde w_{i,t}}{N_{i,t}+1}}\biggr]
\le\E\biggl[\sqrt{\sum_{t,i}\frac{\bar s_i\ind_{G_t}I_{i,t}}{N_{i,t}+1}}\,
\sqrt{\sum_{t,i}\ind_{G_t}I_{i,t}\widetilde w_{i,t}}\biggr]\\
&\le\sqrt{\E\biggl[\sum_{t,i}\frac{\bar s_i\ind_{G_t}I_{i,t}}{N_{i,t}+1}\biggr]
\E\biggl[\sum_{t,i}\ind_{G_t}I_{i,t}\widetilde w_{i,t}\biggr]}
\le\sqrt{S(1+\log T)W},
\end{aligned}
\end{equation}
where the first inequality follows from the Cauchy--Schwarz inequality applied to the sum over $(t,i)$ for each realization, the second follows from the Cauchy--Schwarz inequality applied to the expectation of the product, and the third follows from~\eqref{eq:harmonic-log}, $\ind_{G_t}\le1$, $S=\sum_i \bar s_i$, and $\sum_i\widetilde w_{i,t}I_{i,t}=\sum_hw_{t,h}$.
Let $\Lambda'=\Lambda\log T(1+\log T)$.
Combining~\eqref{eq:local-error}, \eqref{eq:harmonic-log}, and~\eqref{eq:cs-first}, we obtain
\begin{equation}\label{eq:gap-variance}
\begin{aligned}
\Delta
&\le24\sqrt{\Lambda\log T}\,\E\biggl[\sum_{t,i}\ind_{G_t}I_{i,t}\sqrt{\frac{\bar s_i\widetilde w_{i,t}}{N_{i,t}+1}}\biggr]
+417\Lambda\,\E\biggl[\sum_{t,i}\frac{\bar s_i\ind_{G_t}I_{i,t}}{N_{i,t}+1}\biggr]\\
&\le24\sqrt{S\Lambda'W}+417S\Lambda(1+\log T).
\end{aligned}
\end{equation}
By~\eqref{eq:local-budget-averaged}, we have $\sum_t\E[\ind_{G_t}\reg_t]\le\Delta\le24\sqrt{ST\Lambda'}+417S\Lambda(1+\log T)$.
Adding $\delta T$ from~\eqref{eq:regret-good-bad} proves~\eqref{eq:local-theorem} with $c=417$.
\end{proof}

Proposition~\ref{prop:first-order} follows from~\eqref{eq:gap-variance} with a sharper bound on $W$ than~\eqref{eq:local-budget-averaged}.
\begin{proof}[Proof of Proposition~\ref{prop:first-order}]
Fix a history before round $t$ for which $G_t$ holds, and take expectations conditional on it, as in the proof of Theorem~\ref{thm:local}.
Let $\Delta_t=\E[\sum_{h=0}^nd_{t,h}]$, which equals $\E[Y_{t,0}-Y_{t,n+1}]$ by the tower property.
By optimism, $\E[Y_{t,0}^2]\ge(\E[Y_{t,0}])^2\ge p^2$, and since $Y_{t,n+1}=f(\sigma^t,X^t)$, we have $\E[Y_{t,n+1}^2]\le\E[Y_{t,n+1}]=f(\sigma^t\mid D)\le p$.
Since $d_{t,h}\ge0$, we have $(d_{t,h})_+=d_{t,h}$, and hence the first claim of Lemma~\ref{lem:variance-budget} yields
\[
\E\biggl[\sum_{h=0}^nw_{t,h}\biggr]\le p-p^2+2\Delta_t=p(1-p)+2\Delta_t.
\]
Since $\Delta=\sum_t\E[\ind_{G_t}\Delta_t]$, averaging this bound over good histories and summing over $t$ yields $W\le Tp(1-p)+2\Delta$.
Substituting this bound into~\eqref{eq:gap-variance} yields
\[
\Delta\le24\sqrt{S\Lambda'(Tp(1-p)+2\Delta)}+417S\Lambda(1+\log T).
\]
Using $\sqrt{x+y}\le\sqrt{x}+\sqrt{y}$ for $x,y\ge0$ and $24\sqrt{2S\Lambda'\Delta}\le\Delta/2+576S\Lambda'$, we obtain
\begin{align*}
\Delta
&\le24\sqrt{ST\,p(1-p)\Lambda'}+24\sqrt{2S\Lambda'\Delta}+417S\Lambda(1+\log T)\\
&\le24\sqrt{ST\,p(1-p)\Lambda'}+\frac{\Delta}{2}+576S\Lambda'+417S\Lambda(1+\log T).
\end{align*}
Subtracting $\Delta/2$ from both sides and multiplying by $2$ yields
\[
\Delta\le48\sqrt{ST\,p(1-p)\Lambda'}+1152S\Lambda'+834S\Lambda(1+\log T).
\]
Finally, $R_T\le\Delta+\delta T$ by~\eqref{eq:regret-good-bad} and $\sum_t\E[\ind_{G_t}\reg_t]\le\Delta$, and $\delta T<1$ for $\delta=1/(T+1)$.
\end{proof}

Proposition~\ref{prop:single} follows from the same calculation, with the first claim of Lemma~\ref{lem:variance-budget} in place of the second.
\begin{proof}[Proof of Proposition~\ref{prop:single}]
We treat~(a) as the case $\alpha=1$ and bound $R_T^{(\alpha)}$.
Let $G_t$ be the event that $D_i\in\CC_{i,t}$ for every $i$, as in the proof of Theorem~\ref{thm:local}.

Fix a history before round $t$ for which $G_t$ holds, take expectations conditional on it, and let $Q=Q^t$.
If $\sigma^t$ is randomized, we also condition on the deterministic policy drawn from it.
Define the visited nodes $u_{t,h}$, the last stage $h^*$, and the filtration $(\FF_{t,h})_h$ as in the proof of Theorem~\ref{thm:local}.
In~(a), let $Y_{t,h}=g^Q_{u_{t,h}}(X^t_{\obs(u_{t,h})})$.
In~(b), let $Y_{t,h}$ be the expected reward of continuing the deterministic policy from $u_{t,h}$ when the unobserved coordinates follow $Q$.
In both cases, $Y_{t,h}\in[0,1]$ and $Y_{t,h}=f_{\ell^t}(X^t_{\obs(\ell^t)})$ for $h^*\le h\le n+1$.
Define $d_{t,h}$ and $w_{t,h}$ as in the proof of Theorem~\ref{thm:local}.

Let $h<h^*$.
If the chosen edge is a stopping edge, then $Y_{t,h+1}=Y_{t,h}$, in~(a) since the chosen child maximizes $h^Q$ and in~(b) by the definition of $Y_{t,h}$.
This equality also holds for $h^*\le h\le n$, and hence $d_{t,h}=w_{t,h}=0$ in these cases.
If the chosen edge $(u,v)$ reveals coordinate $i$, let $\phi(z)$ be the value at $v$ for $x_i=z$, which is $g^Q_v(z,x_{\obs(u)})$ in~(a).
Then, $Y_{t,h}=\E_{X\sim Q_i}[\phi(X)]$ for the same reasons, whereas $Y_{t,h+1}=\phi(X_i^t)$ and $X_i^t$ follows $D_i$ conditional on $\FF_{t,h}$.
Hence,
\[
d_{t,h}=\E_{X\sim Q_i}[\phi(X)]-\E_{X\sim D_i}[\phi(X)],
\qquad
w_{t,h}=\Var_{X\sim D_i}(\phi(X)).
\]
Unlike in the proof of Theorem~\ref{thm:local}, $d_{t,h}$ can be negative, since $Q_i$ need not maximize the expectation of $\phi$ over $\CC_{i,t}$, and hence $(Y_{t,h})_h$ need not be a supermartingale.

We bound the regret in this round by the positive parts of the gaps.
Since $\obs(r)=\emptyset$, the expectation of $Y_{t,0}$ over the randomness of $\sigma^t$ is $f(\sigma^t\mid Q)$, which is at least $\alpha\,\opt(Q)\ge\alpha\,\opt(D)-1/T$ since $D_i\in\CC_{i,t}$ for every $i$.
Taking expectations also over this randomness, the tower property yields $\E[Y_{t,0}]-\E[f(\sigma^t,X^t)]=\E[\sum_{h=0}^nd_{t,h}]$, and hence
\begin{equation}\label{eq:single-round}
\alpha\,\opt(D)-\E[f(\sigma^t,X^t)]\le\frac1T+\E\biggl[\sum_{h=0}^n(d_{t,h})_+\biggr].
\end{equation}
Since $Y_{t,h}\in[0,1]$, the first claim of Lemma~\ref{lem:variance-budget} yields
\begin{equation}\label{eq:single-budget}
\E\biggl[\sum_{h=0}^nw_{t,h}\biggr]
\le\E[Y_{t,n+1}^2]-\E[Y_{t,0}^2]+2\E\biggl[\sum_{h=0}^n(d_{t,h})_+\biggr]
\le1+2\E\biggl[\sum_{h=0}^n(d_{t,h})_+\biggr].
\end{equation}

We next bound each positive part, using $Q_i,D_i\in\CC_{i,t}$ and $m=N_{i,t}$.
In~(a), if $s_i\le k_i$, then $\phi$ has level-set complexity at most $s_i$ by the condition in~(a), and Lemma~\ref{lem:level-stability}(b) with $P=D_i$ applies.
Otherwise, Lemma~\ref{lem:finite-support} with $k=k_i$ and $P=D_i$ applies.
In~(b), Lemma~\ref{lem:finite-support} with $k=k_i$ and $P=D_i$ applies to every $\phi$.
Using $\log(N_{i,t}+1)\le\log T$, we obtain~\eqref{eq:local-error} with $(d_{t,h})_+$ in place of $d_{t,h}$, where $\bar s_i=\min\{s_i,k_i\}$ in~(a) and $\bar s_i=k_i$ in~(b).

Finally, we sum over the rounds, with expectations also averaging over the histories.
Let $\Delta$ and $W$ be the sums over $t$ of $\E[\ind_{G_t}\sum_h(d_{t,h})_+]$ and $\E[\ind_{G_t}\sum_hw_{t,h}]$, respectively.
By~\eqref{eq:single-budget}, we have $W\le T+2\Delta$.
The calculation in~\eqref{eq:cs-first} and~\eqref{eq:gap-variance}, with $(d_{t,h})_+$ in place of $d_{t,h}$, yields $\Delta\le24\sqrt{S\Lambda'(T+2\Delta)}+417S\Lambda(1+\log T)$, with $K$ in place of $S$ in~(b).
Using $\sqrt{x+y}\le\sqrt{x}+\sqrt{y}$ and $24\sqrt{2S\Lambda'\Delta}\le\Delta/2+576S\Lambda'$ as in the proof of Proposition~\ref{prop:first-order}, we obtain $\Delta\le48\sqrt{ST\Lambda'}+1152S\Lambda'+834S\Lambda(1+\log T)$.
By~\eqref{eq:single-round}, the good rounds contribute at most
$1+\Delta$ to $R_T^{(\alpha)}$.
Each other round contributes at most 1 and occurs with probability
at most $\delta$ by Lemma~\ref{lem:coverage}.
Hence, $R_T^{(\alpha)}\le1+\Delta+\delta T$.
\end{proof}

\subsection{Extension to Immediate Rewards}\label{subsec:sums}
The proof of Theorem~\ref{thm:local} in Section~\ref{subsec:main-upper} uses Assumption~\ref{ass:level-set} only in the local error bound~\eqref{eq:local-error}, and we prove Theorem~\ref{thm:sum} by replacing this bound with one that treats the immediate reward and the residual value separately.
\begin{proof}[Proof of Theorem~\ref{thm:sum}]
Consider a round in which $G_t$ holds and a stage $h$ at which the chosen edge $e=(u,v)$ reveals coordinate $i$ after observations $x_{\obs(u)}$.
Let $\kappa=\kappa_e(x_{\obs(u)})$ and $F_\ell=F_{e,\ell}$ be as in condition~(a) of Theorem~\ref{thm:sum}, and write $\chi(z)=\chi_e(z,x_{\obs(u)})$.
Let $\phi(z)$ be the optimistic continuation value at $v$ for $x_i=z$, defined by~\eqref{eq:optimistic-dp} and~\eqref{eq:optimistic-edge} on the subtree of $v$ with $F_\ell$ in place of $f_\ell$.
Since $\kappa+\chi(x_i)$ is determined by the observations at $v$, and expectations, suprema, and maxima commute with adding a constant, we have $\bar g_v^{\,t}(z,x_{\obs(u)})=\kappa+\chi(z)+\phi(z)$.
The function $\phi$ takes values in $[0,1]$ since every $F_\ell$ does, and it is nonincreasing by condition~(b), since expectations, suprema, and maxima preserve monotonicity.
Hence, $\phi$ has level-set complexity at most $1\le s_i$.

Let $w_{t,h}$ be as in Section~\ref{subsec:main-upper}, let $\sigma_\phi$ and $\sigma_\chi$ be the standard deviations of $\phi(X_i)$ and $\chi(X_i)$ for $X_i\sim D_i$, and let $\nu_{t,h}=\E_{X_i\sim D_i}[\chi(X_i)]$.
Since $\phi=\bar g_v^{\,t}-\kappa-\chi$, the triangle inequality for standard deviations implies $\sigma_\phi\le\sqrt{w_{t,h}}+\sigma_\chi$.
Moreover, $\sigma_\chi^2\le\nu_{t,h}$ since $\chi$ takes values in $[0,1]$.
Combining these bounds with the arithmetic-quadratic mean inequality applied to $\sqrt{w_{t,h}}$, $\sqrt{\nu_{t,h}}$, and $\sqrt{\nu_{t,h}}$, we obtain
\[
\sigma_\phi+\sigma_\chi\le\sqrt{w_{t,h}}+2\sqrt{\nu_{t,h}}\le\sqrt{3(w_{t,h}+2\nu_{t,h})}.
\]
Since $\kappa$ does not depend on $X_i$, it cancels in $d_{t,h}$, and hence
\[
d_{t,h}
\le\sup_{Q\in\CC_{i,t}}\bigl(\E_{X_i\sim Q}[\chi(X_i)]-\E_{X_i\sim D_i}[\chi(X_i)]\bigr)
+\sup_{Q\in\CC_{i,t}}\bigl(\E_{X_i\sim Q}[\phi(X_i)]-\E_{X_i\sim D_i}[\phi(X_i)]\bigr).
\]
Applying Lemma~\ref{lem:level-stability}(b) to $\chi$ and $\phi$ with $m=N_{i,t}$ and $s=s_i$, and then the preceding bound on $\sigma_\chi+\sigma_\phi$, we obtain
\begin{equation}\label{eq:sum-local}
0\le d_{t,h}
\le24\sqrt{\frac{s_i\Lambda\log T}{N_{i,t}+1}}\,(\sigma_\chi+\sigma_\phi)+\frac{834s_i\Lambda}{N_{i,t}+1}
\le24\sqrt{\frac{3s_i\Lambda\log T}{N_{i,t}+1}\bigl(w_{t,h}+2\nu_{t,h}\bigr)}+\frac{834s_i\Lambda}{N_{i,t}+1}.
\end{equation}
If $D_i$ has $k_i<s_i$ support points, then Lemma~\ref{lem:finite-support} for $\chi$ with $k=k_i$ and Lemma~\ref{lem:level-stability}(b) for $\phi$ with $s=1\le k_i$ yield~\eqref{eq:sum-local} with $k_i$ in place of $s_i$, which proves the remark after Theorem~\ref{thm:sum}.

At a stage that reveals no coordinate, set $\nu_{t,h}=0$.
Then, $\nu_{t,h}$ is the conditional expectation, given $\FF_{t,h}$, of the immediate reward collected at stage $h$.
By condition~(c), these immediate rewards sum to at most 1 along the executed path, and hence $\E[\sum_h\nu_{t,h}]\le1$.
Together with~\eqref{eq:local-round-budget}, we obtain $\E[\sum_h(w_{t,h}+2\nu_{t,h})]\le3$.
Define $\widetilde\nu_{i,t}$ from $\nu_{t,h}$ in the same way as $\widetilde w_{i,t}$ from $w_{t,h}$.
Let $\Lambda'=\Lambda\log T(1+\log T)$, $\Delta=\sum_t\E[\ind_{G_t}\sum_hd_{t,h}]$, and $W'=\sum_t\E[\ind_{G_t}\sum_h(w_{t,h}+2\nu_{t,h})]$, which is at most $3T$ by the preceding bound.
The calculation in~\eqref{eq:cs-first} and~\eqref{eq:gap-variance}, with~\eqref{eq:sum-local} in place of~\eqref{eq:local-error}, $s_i$ in place of $\bar s_i$, and $\widetilde w_{i,t}+2\widetilde\nu_{i,t}$ in place of $\widetilde w_{i,t}$, yields
\begin{equation}\label{eq:sum-gap-variance}
\Delta\le24\sqrt{3S\Lambda'W'}+834S\Lambda(1+\log T).
\end{equation}
Since $W'\le3T$, we have $\sum_t\E[\ind_{G_t}\reg_t]\le\Delta\le72\sqrt{ST\Lambda'}+834S\Lambda(1+\log T)$, and adding $\delta T$ from~\eqref{eq:regret-good-bad} proves the claim with $c=834$.
\end{proof}

Proposition~\ref{prop:sum-first-order} follows from~\eqref{eq:sum-gap-variance} with a sharper bound on $W'$ than $W'\le3T$.
\begin{proof}[Proof of Proposition~\ref{prop:sum-first-order}]
Fix a history before round $t$ for which $G_t$ holds, and let $\Delta_t$ be as in the proof of Proposition~\ref{prop:first-order}, which shows $\E[\sum_hw_{t,h}]\le p(1-p)+2\Delta_t$ without using Assumption~\ref{ass:level-set}.
By the additional condition, the immediate rewards along the executed path sum to at most $f(\sigma^t,X^t)$, and hence $\E[\sum_h\nu_{t,h}]\le f(\sigma^t\mid D)\le p$.
Thus, $\E[\sum_h(w_{t,h}+2\nu_{t,h})]\le3p+2\Delta_t$.
Averaging over good histories and summing over $t$ yields $W'\le3Tp+2\Delta$, and substituting this bound into~\eqref{eq:sum-gap-variance} yields
\[
\Delta\le24\sqrt{3S\Lambda'(3Tp+2\Delta)}+834S\Lambda(1+\log T).
\]
Absorbing the term proportional to $\sqrt\Delta$ as in the proof of Proposition~\ref{prop:first-order} and using $R_T\le\Delta+\delta T$ yields $R_T=\otil(\sqrt{STp}+S)$.
\end{proof}

\section{Lower Bound}\label{sec:lower}
We prove Theorem~\ref{thm:lower} with an instance in which each coordinate $i$ differs from a special label with probability $s_i/(2S)$ and the learner must predict the signs carried by these rare values.
The use of rare values is inspired by a lower-bound construction of Guo, Huang, Tang, and Zhang~\cite[Theorem~7]{pmlr-v134-guo21a} for the sample complexity of learning over product distributions.
We embed the instance into a decision tree and order the labels such that the level-set complexity of coordinate $i$ is at most $s_i$.

\subsection{Construction}\label{subsec:lower-construction}
Fix integers $T,s_1,\ldots,s_n\ge1$, let $S=s_1+\cdots+s_n$, and set
\begin{equation}\label{eq:lower-parameters}
\varepsilon=\frac14\min\left\{1,\;\sqrt{\frac{S}{T}}\right\}.
\end{equation}
Let $J=\{(i,j):i\in[n],\ j\in[s_i]\}$, which has $S$ elements.
An unknown sign vector $\boldsymbol\theta=(\theta_{ij})_{(i,j)\in J}\in\{-1,+1\}^J$ is fixed throughout all $T$ rounds.

\paragraph{Coordinates and distributions.}
Coordinate $i$ takes values in a set of $2s_i+1$ labels, which consists of a special label $\bot$ and the pairs $(j,y)\in[s_i]\times\{-1,+1\}$.
Its distribution is
\begin{equation*}
D_i^{\boldsymbol\theta}(\bot)=1-\frac{s_i}{2S},\qquad
D_i^{\boldsymbol\theta}(j,y)=\frac{1+y\varepsilon\theta_{ij}}{4S}.
\end{equation*}
Equivalently, $X_i=\bot$ with probability $1-s_i/(2S)$, and otherwise $X_i=(j,y)$, where the column $j$ is drawn uniformly from $[s_i]$ and then the sign $y$ is $\theta_{ij}$ with probability $(1+\varepsilon)/2$ and $-\theta_{ij}$ otherwise.
Since $\varepsilon\le1/4$, all $2s_i+1$ labels have positive probability.
Let $D^{\boldsymbol\theta}=\prod_{i=1}^nD_i^{\boldsymbol\theta}$.
We identify the labels
\[
(1,-1),\;\bot,\;(1,+1),\;(2,-1),\;(2,+1),\;\ldots,\;(s_i,-1),\;(s_i,+1)
\]
with the points $0,1/(2s_i),2/(2s_i),\ldots,1$ of $[0,1]$, respectively.
The label $\bot$ thus corresponds to the point $1/(2s_i)$, and only the order of the points matters for level-set complexity.
Placing $\bot$ between the two labels of column 1 keeps the level-set complexity of coordinate $i$ at most $s_i$, as we verify after describing the rewards.

\paragraph{Decision tree and rewards.}
There are no initial observations.
At the root, the learner chooses a prediction vector $\widehat{\boldsymbol y}=(\widehat y_{ij})_{(i,j)\in J}\in\{-1,+1\}^J$, where $\widehat y_{ij}$ predicts the sign $y$ when $X_i=(j,y)$.
The edge chosen at the root reveals $X_1$, and a chain of edges then reveals $X_2,\ldots,X_n$ in order, where each intermediate node has a single child and the edge revealing $X_n$ leads to a leaf (for $n=1$, this is the edge chosen at the root).
For a realization $x=(x_1,\ldots,x_n)$, the reward $f_{\widehat{\boldsymbol y}}(x)$ at the leaf corresponding to $\widehat{\boldsymbol y}$ is 1 if exactly one coordinate $i$ satisfies $x_i\ne\bot$ and this coordinate satisfies $x_i=(j,\widehat y_{ij})$ for some $j\in[s_i]$, and 0 otherwise.
Thus, every round reveals all coordinates regardless of the prediction, and only the distributions depend on $\boldsymbol\theta$.

\paragraph{Level-set complexity.}
We verify Assumption~\ref{ass:level-set} with $s_1,\ldots,s_n$.
Fix nonempty sets of distributions, a prediction vector $\widehat{\boldsymbol y}$, a coordinate $i$, and values $x_1,\ldots,x_{i-1}$, and let $\phi(z)$ be the optimistic continuation value after $X_i=z$ is revealed.
If $z=(j,-\widehat y_{ij})$ for some $j\in[s_i]$, then coordinate $i$ differs from $\bot$ and its sign is predicted incorrectly, and hence $\phi(z)=0$.
Since $\phi\ge0$, every superlevel set of $\phi$ at a negative level is the whole domain, and every superlevel set at a nonnegative level is a subset of $\{\bot\}\cup\{(j,\widehat y_{ij}):j\in[s_i]\}$.
Such a subset consists of at most $s_i$ runs of consecutive labels, since it contains at most one label from each column and $\bot$ is adjacent to both labels of column 1.
Since a run of consecutive labels is the intersection of the domain with an interval, $\phi$ has level-set complexity at most $s_i$.

The next lemma shows that each incorrect sign in $\widehat{\boldsymbol y}$ reduces the expected reward by at least a fixed amount.
\begin{lemma}\label{lem:prediction-gap}
For every $\boldsymbol\theta$ and every prediction vector $\widehat{\boldsymbol y}$,
\begin{equation*}
\opt(D^{\boldsymbol\theta})-\E_{X\sim D^{\boldsymbol\theta}}[f_{\widehat{\boldsymbol y}}(X)]
\ge\frac{\varepsilon}{4S}\,\bigl|\{(i,j)\in J:\widehat y_{ij}\ne\theta_{ij}\}\bigr|.
\end{equation*}
\end{lemma}
\begin{proof}
Let $X\sim D^{\boldsymbol\theta}$, and let $c_i$ be the probability that $X_k=\bot$ for every $k\ne i$.
By the union bound, $c_i\ge1-\sum_{k\ne i}s_k/(2S)\ge1/2$.
For $(i,j)\in J$, the event that $X_i=(j,\widehat y_{ij})$ and $X_k=\bot$ for every $k\ne i$ has probability $c_i(1+\varepsilon\widehat y_{ij}\theta_{ij})/(4S)$.
These events are disjoint, and $f_{\widehat{\boldsymbol y}}(X)=1$ exactly when one of them occurs.
Hence, $\E[f_{\widehat{\boldsymbol y}}(X)]=\sum_{(i,j)\in J}c_i(1+\varepsilon\widehat y_{ij}\theta_{ij})/(4S)$, and each incorrect sign decreases this value by $c_i\varepsilon/(2S)\ge\varepsilon/(4S)$ relative to $\widehat{\boldsymbol y}=\boldsymbol\theta$.
Since the root is the only decision and randomizing over root actions averages their values, $\opt(D^{\boldsymbol\theta})$ is the expected reward for $\widehat{\boldsymbol y}=\boldsymbol\theta$, which proves the claim.
\end{proof}

\subsection{Proof of the Lower Bound}\label{subsec:lower-proof}
Before proving Theorem~\ref{thm:lower}, we show that every learning algorithm predicts each sign $\theta_{ij}$ incorrectly with constant probability when $\boldsymbol\theta$ is chosen at random.
For distributions $P,Q$ on a finite set $\mathcal Z$, we write $\KL(P\|Q)=\sum_{z\in\mathcal Z}P(z)\log(P(z)/Q(z))$ for the KL divergence and $\TV(P,Q)=\max_{A\subseteq\mathcal Z}|P(A)-Q(A)|$ for the total variation distance.
Fix any learning algorithm, and let $\widehat{\boldsymbol y}^t$ be its prediction vector in round $t\in[T]$.
Choose $\boldsymbol\theta$ uniformly from $\{-1,+1\}^J$ once at the beginning, and let $\overline{\Prb}$ denote the probability over $\boldsymbol\theta$, the observations, and the internal randomness of the learning algorithm.
\begin{lemma}\label{lem:testing-lower}
For every $t\in[T]$ and $(i,j)\in J$, we have $\overline{\Prb}(\widehat y_{ij}^t\ne\theta_{ij})\ge3/8$.
\end{lemma}
\begin{proof}
Fix $t$, $(i,j)$, and the signs $\theta_{i'j'}$ for all $(i',j')\in J\setminus\{(i,j)\}$.
Let $\mathcal Z$ be the finite set of possible values of the past observations $Z=(X_k^\tau)_{\tau<t,k\in[n]}$, and let $P_+,P_-$ be the distributions of $Z$ when $\theta_{ij}=+1$ and $\theta_{ij}=-1$, respectively.
Since every round reveals all coordinates regardless of the predictions, $P_+$ and $P_-$ do not depend on the learning algorithm.
For one observation of coordinate $i$, only the probabilities of the labels $(j,-1)$ and $(j,+1)$ differ, and their contributions to the KL divergence sum to $\varepsilon\log((1+\varepsilon)/(1-\varepsilon))/(2S)$.
By independence across coordinates and rounds, we have
\begin{equation*}
\KL(P_+\|P_-)
=\frac{(t-1)\varepsilon}{2S}\log\frac{1+\varepsilon}{1-\varepsilon}
\le\frac{3T\varepsilon^2}{2S}\le\frac3{32},
\end{equation*}
where we used $\log((1+\varepsilon)/(1-\varepsilon))\le3\varepsilon$, which holds for $\varepsilon\le1/4$ because the left-hand side vanishes at $\varepsilon=0$ and its derivative $2/(1-\varepsilon^2)$ is at most 3, and then $\varepsilon\le\sqrt{S/T}/4$.

By Pinsker's inequality~\cite{lattimore2020bandit}, $\TV(P_+,P_-)\le\sqrt{\KL(P_+\|P_-)/2}\le\sqrt{3/64}<1/4$.
Since the learner's internal randomness is independent of the observations, the conditional probability $a(Z)$ that $\widehat y_{ij}^t=+1$ given $Z$ is the same $[0,1]$-valued function of $Z$ in both environments.
Let $A=\{z\in\mathcal Z:P_+(z)>P_-(z)\}$.
Since $a$ takes values in $[0,1]$, we have
\begin{equation*}
\E_{Z\sim P_+}[a(Z)]-\E_{Z\sim P_-}[a(Z)]
=\sum_{z\in\mathcal Z}a(z)\bigl(P_+(z)-P_-(z)\bigr)
\le P_+(A)-P_-(A)\le\TV(P_+,P_-).
\end{equation*}
Therefore, the error probability averaged over the two values of $\theta_{ij}$ is
\begin{align*}
\frac12\bigl(\E_{Z\sim P_+}[1-a(Z)]+\E_{Z\sim P_-}[a(Z)]\bigr)
&=\frac12\bigl(1-\E_{Z\sim P_+}[a(Z)]+\E_{Z\sim P_-}[a(Z)]\bigr)\\
&\ge\frac12\bigl(1-\TV(P_+,P_-)\bigr)\ge\frac38.
\end{align*}
This bound holds for every choice of the fixed signs $\theta_{i'j'}$, and averaging over them yields $\overline{\Prb}(\widehat y_{ij}^t\ne\theta_{ij})\ge3/8$.
\end{proof}

We are now ready to prove Theorem~\ref{thm:lower}.
\begin{proof}[Proof of Theorem~\ref{thm:lower}]
Consider the problem in Section~\ref{subsec:lower-construction}, which satisfies Assumption~\ref{ass:level-set} with $s_1,\ldots,s_n$.
Conditional on $\boldsymbol\theta$, the past observations, and the learner's randomness, the prediction $\widehat{\boldsymbol y}^t$ is fixed and independent of $X^t\sim D^{\boldsymbol\theta}$.
Hence, $\reg_t=\opt(D^{\boldsymbol\theta})-\E_{X\sim D^{\boldsymbol\theta}}[f_{\widehat{\boldsymbol y}^t}(X)]$, which is at least $\frac{\varepsilon}{4S}|\{(i,j)\in J:\widehat y_{ij}^t\ne\theta_{ij}\}|$ by Lemma~\ref{lem:prediction-gap}.
Taking expectations, summing over $t$, and averaging over $\boldsymbol\theta$, we obtain from Lemma~\ref{lem:testing-lower}
\begin{align*}
2^{-S}\sum_{\boldsymbol\theta\in\{-1,+1\}^J}R_T(D^{\boldsymbol\theta})
&\ge\frac{\varepsilon}{4S}
\sum_{t=1}^T\sum_{(i,j)\in J}
\overline{\Prb}(\widehat y_{ij}^t\ne\theta_{ij})
\ge\frac{\varepsilon}{4S}\cdot TS\cdot\frac38
=\frac{3}{32}T\varepsilon
\ge\frac1{64}\min\{T,\;\sqrt{ST}\},
\end{align*}
where the last inequality uses~\eqref{eq:lower-parameters}.
Hence, $R_T(D^{\boldsymbol\theta})$ is at least this value for some $\boldsymbol\theta\in\{-1,+1\}^J$.
Since all $n$ coordinates are observed in every round and these observations determine the reward of every action, the lower bound holds even under full feedback.
\end{proof}

\section{Conclusion}\label{sec:conclusion}
We studied SOO under semi-bandit feedback without FSD monotonicity and proposed a learning algorithm based on optimistic DP\@.
Its regret is $\otil(\sqrt{ST})$ with $S=\sum_i\min\{s_i,k_i\}$, where $s_i$ bounds the level-set complexity of the optimistic continuation values for coordinate $i$ and $k_i$ is the number of support points of $D_i$, neither of which the algorithm needs to know.
We also showed that this bound is tight up to logarithmic factors in the worst case for all positive integers $s_1,\ldots,s_n$ with $k_i=2s_i+1$, even under full feedback.
With the extension in Theorem~\ref{thm:sum}, the analysis yields regret $\otil(\sqrt{nT})$ for all five applications in Section~\ref{sec:applications} after rescaling the rewards to $[0,1]$, provided that, in scheduling, the reward functions have bounded level-set complexity and the observed processing time and reward identify the outcome.

Three directions remain open.
First, the running time of each round is polynomial in the size of the decision tree given the suprema of the optimistic continuation values over intervals (Proposition~\ref{prop:computation}), but the tree can be exponentially large in $n$, as in SBFE with general Boolean functions, and computing the suprema requires structure such as finite domains, quasi-convexity, or Lipschitz continuity (Appendix~\ref{subsec:states}).
Proposition~\ref{prop:states} covers problems with few reachable states, and FSD-monotone problems in a known direction with finite domains reduce to exact or $\alpha$-approximation algorithms for known distributions by Proposition~\ref{prop:single}, but it is open which other problems admit efficient learning algorithms.
Second, level-set complexity is defined for real-valued coordinates, and it is open how to extend the analysis to multi-dimensional observations.
Third, our algorithm builds confidence sets from the observed values, and it is open whether regret growing as $\sqrt T$ and polynomially in $n$ is possible without FSD monotonicity under bandit feedback, where only the reward is observed, as studied for the prophet inequality and Pandora's box~\cite{Gatmiry2024}.

\section*{Acknowledgments}
This work was partially supported by the joint project of Kyoto University and Toyota Motor Corporation, titled ``Advanced Mathematical Science for Mobility Society'', JST ERATO Grant Number JPMJER2301, JST ASPIRE Grant Number JPMJAP2302, and JSPS KAKENHI Grant Numbers JP21K17708, JP21H03397, JP25K00137, JP26K14718, JP26K02867, and JP26KJ1106.

\section*{AI Methodology}
We wrote earlier versions of this paper without AI assistance.
These versions used the total variation of the functions $\phi(z)=\bar g_v^{\,t}(z,x_{\obs(u)})$ on intervals $I$ for revelation edges $(u,v)$ as a parameter.
Here, the total variation of $\phi$ on $I$ is the supremum of $\sum_{j=0}^{k-1}|\phi(z_{j+1})-\phi(z_j)|$ over all $k\ge0$ and $z_0<\cdots<z_k$ in $I\cap\operatorname{dom}(\phi)$.
An analysis based on this parameter yields regret $\otil(n^{3/4}\sqrt{\nu T})$ when the total variation is at most $\nu$.
In the current version, the idea of using level-set complexity and the proof of the lower bound came from OpenAI ChatGPT/Codex and Anthropic Claude.
The tools materially affected Sections~\ref{subsec:level-set}, \ref{sec:algorithm}, \ref{sec:applications}, \ref{sec:analysis}, and~\ref{sec:lower} and Appendix~\ref{app:lp}.
The authors verified the correctness and originality of all content, including references.
We also used these tools to assist with strengthening the content and editing the language throughout the paper.

\appendix
\section{Computational Aspects}\label{app:lp}
We reduce each optimistic update to an LP, bound the running time of one round given the suprema of optimistic continuation values over intervals, and compute these suprema when the observations can be summarized by few reachable states.

\subsection{Finite Reduction to Interval Masses}\label{subsec:interval-masses}
Fix a coordinate $i$, an edge $(u,v)$, and observations $x_{\obs(u)}$, and let $\phi(z)=\bar g_v^{\,t}(z,x_{\obs(u)})$.
The optimistic update $\bar h_{(u,v)}^{\,t}(x_{\obs(u)})=\sup_{Q\in\CC_{i,t}}\E_{X\sim Q}[\phi(X)]$ in~\eqref{eq:optimistic-edge} maximizes over all distributions $Q$ on the domain of coordinate $i$ that satisfy the constraint in~\eqref{eq:confidence} for every interval $I\in\II_1$, and hence, in general, over an infinite-dimensional set with infinitely many constraints.
We reduce it to an LP over the masses of finitely many cells, which are formed by the sample points and the intervals between them, given the suprema of $\phi$ on the cells.
If $N_{i,t}<16$, the confidence set consists of all distributions on that coordinate's domain, by definition if $N_{i,t}=0$ and since $b_{N_{i,t}}\equiv1$ as in the proof of Lemma~\ref{lem:coverage} otherwise, and the optimistic update equals $\sup_{z\in\operatorname{dom}(\phi)}\phi(z)$.
Otherwise, write $N=N_{i,t}\ge16$, and let $\widehat D_{i,N}$ be the empirical distribution of coordinate $i$, as in Section~\ref{subsec:optimistic-dp}.

Let $y_1<\cdots<y_{m'}$ be the distinct sample points of the empirical distribution.
Partition $[0,1]$ into the singletons $\{y_j\}$, the open intervals between them, and the two end intervals $[0,y_1)$ and $(y_{m'},1]$, and intersect each set with $\operatorname{dom}(\phi)$.
List the nonempty sets from left to right as $C_1,\ldots,C_M$, which we call \emph{cells}.
We have $M\le2m'+1\le2N+1$.
Let $N_j=N\cdot\widehat D_{i,N}(C_j)$ be the number of samples in $C_j$, and let $\bar\phi_j=\sup_{z\in C_j}\phi(z)$.
Each $N_j$ is a nonnegative integer, and $\sum_{j=1}^MN_j=N$.
A cell containing a sample point is the singleton consisting of that point.

Writing $q_j$ for the mass assigned to cell $C_j$, define
\begin{equation}\label{eq:cell-constraints}
\mathcal Q=\biggl\{q\in\mathbb R_{\ge0}^M:\sum_{j=1}^Mq_j=1,\ \biggl|\sum_{j=a}^cq_j-\frac1N\sum_{j=a}^cN_j\biggr|\le b_N\biggl(\frac1N\sum_{j=a}^cN_j\biggr)\text{ for all }1\le a\le c\le M\biggr\}.
\end{equation}
We write $q_{\le c}=q_1+\cdots+q_c$ for the cumulative masses, with $q_{\le0}=0$.

\begin{proposition}\label{prop:local-lp}
The set $\mathcal Q$ is nonempty and compact, and the optimistic update satisfies
\begin{equation}\label{eq:cell-objective}
\sup_{Q\in\CC_{i,t}}\E_{X\sim Q}[\phi(X)]
=\max_{q\in\mathcal Q}\sum_{j=1}^M q_j\bar\phi_j.
\end{equation}
\end{proposition}
\begin{proof}
For any $Q\in\CC_{i,t}$, let $q_j=Q(C_j)$.
For $1\le a\le c\le M$, there is an interval $I\in\II_1$ with $I\cap\operatorname{dom}(\phi)=C_a\cup\cdots\cup C_c$, and hence $Q(I)=\sum_{j=a}^cq_j$ and $\widehat D_{i,N}(I)=\frac1N\sum_{j=a}^cN_j$.
Thus, the constraint in~\eqref{eq:confidence} for this interval $I$, which $Q$ satisfies, is the constraint in~\eqref{eq:cell-constraints} for $(a,c)$, and hence $q\in\mathcal Q$.
In addition, $\E_{X\sim Q}[\phi(X)]\le\sum_{j=1}^Mq_j\bar\phi_j$.
Since the empirical distribution $\widehat D_{i,N}$ belongs to $\CC_{i,t}$, this argument shows that the vector with $q_j=\widehat D_{i,N}(C_j)$ lies in $\mathcal Q$.
As a closed subset of the simplex, $\mathcal Q$ is therefore nonempty and compact, and the linear objective $\sum_{j=1}^Mq_j\bar\phi_j$ with coefficients $\bar\phi_j\in[0,1]$ attains its maximum on $\mathcal Q$.
Hence, the left-hand side of~\eqref{eq:cell-objective} is at most the right-hand side.

Conversely, fix $q\in\mathcal Q$.
Choose any point $z_j\in C_j$ from each cell, and let $Q_q$ place mass $q_j$ at $z_j$.
For any $I\in\II_1$, the set $\{j:z_j\in I\}$ consists of consecutive indices $a,\ldots,c$, since the $z_j$ are ordered.
Since sample points are preserved as singleton cells, we also have $\widehat D_{i,N}(I)=\frac1N\sum_{j=a}^cN_j$, and the constraints in~\eqref{eq:cell-constraints} imply $|Q_q(I)-\widehat D_{i,N}(I)|\le b_N(\widehat D_{i,N}(I))$.
If no $z_j$ lies in $I$, both probabilities are 0.
Thus, $Q_q\in\CC_{i,t}$.

Let $q$ be a maximizing vector on the right-hand side.
For any $\zeta>0$, choose a point $z_j$ in each cell $C_j$ such that $\phi(z_j)\ge\bar\phi_j-\zeta$, which exists by the definition of supremum.
Let $Q_q$ place mass $q_j$ at these points $z_j$.
Since the previous paragraph applies to any choice of points in the cells, $Q_q\in\CC_{i,t}$, and hence the left-hand side is at least $\E_{X\sim Q_q}[\phi(X)]=\sum_{j=1}^Mq_j\phi(z_j)\ge\sum_{j=1}^Mq_j\bar\phi_j-\zeta$.
Equality follows by letting $\zeta\downarrow0$.
\end{proof}

Section~\ref{subsec:regret} uses the following consequence of Proposition~\ref{prop:local-lp} for FSD-monotone problems.
\begin{lemma}\label{lem:dominating}
If the domain of coordinate $i$ is finite, then $\CC_{i,t}$ contains a distribution that dominates all distributions in $\CC_{i,t}$ and a distribution that is dominated by all distributions in $\CC_{i,t}$, and both can be computed by solving an LP.
\end{lemma}
\begin{proof}
If $N_{i,t}<16$, then $\CC_{i,t}$ consists of all distributions on the domain, and the point masses at the largest and the smallest points of the domain suffice.
Otherwise, let the cells $C_1,\ldots,C_M$ and the set $\mathcal Q$ be as above, formed in the domain of coordinate $i$.
Every $q\in\mathcal Q$ satisfies $q_{\le M}=1$, and each constraint in~\eqref{eq:cell-constraints} and each constraint $q_j\ge0$ bounds a difference $q_{\le c}-q_{\le a-1}$ with $1\le a\le c\le M$ from above or below by a constant.
Such constraints are preserved under componentwise minima of the cumulative masses, and hence, for $q,q'\in\mathcal Q$, the vector whose cumulative masses are $\min\{q_{\le c},q'_{\le c}\}$ also belongs to $\mathcal Q$.
Since $\mathcal Q$ is compact, each $q_{\le c}$ attains its minimum on $\mathcal Q$, and applying this closure to the minimizers for $c=1,\ldots,M$ yields $\underline q\in\mathcal Q$ with $\underline q_{\le c}=\min_{q\in\mathcal Q}q_{\le c}$ for every $c$.
This vector is the unique optimal solution of the LP that minimizes $\sum_{c=1}^Mq_{\le c}$ over $\mathcal Q$.
Let $Q^+$ place mass $\underline q_j$ at the largest point of $C_j$, which belongs to $\CC_{i,t}$ by the proof of Proposition~\ref{prop:local-lp}.
For every $Q\in\CC_{i,t}$, the vector $(Q(C_j))_j$ belongs to $\mathcal Q$ by the same proof.
For every $x\in[0,1]$, the cells contained in $[0,x]$ are $C_1,\ldots,C_c$ for some $c$, and hence $Q^+([0,x])=\underline q_{\le c}\le\sum_{j\le c}Q(C_j)\le Q([0,x])$, that is, $Q^+$ dominates $Q$.
Symmetrically, the vector whose cumulative sums are all the largest, with its masses placed at the smallest points of the cells, yields a distribution in $\CC_{i,t}$ that is dominated by all distributions in $\CC_{i,t}$.
\end{proof}

\subsection{Running Time with the Interval-Supremum Oracle}\label{subsec:computation}
By Proposition~\ref{prop:local-lp}, each optimistic update reduces to the LP on the right-hand side of~\eqref{eq:cell-objective}, and it remains to solve this LP and to obtain its coefficients $\bar\phi_j$.
We bound the running time of one round of Algorithm~\ref{alg:main} on the decision tree $(V,A)$ under the following assumptions.
First, we use the real-arithmetic model, in which arithmetic operations, comparisons, square roots, and logarithms have unit cost.
Second, the algorithm can evaluate leaf rewards $f_\ell(x_{\obs(\ell)})$, test feasibility $v\in\mathcal A_u(x_{\obs(u)})$, and call an \emph{interval-supremum oracle}, and each of these operations has unit cost.
Given an edge $(u,v)$ revealing a coordinate $i$, observations $x_{\obs(u)}$, and an interval $I\in\II_1$, the interval-supremum oracle returns $\sup_{z\in I\cap\operatorname{dom}(\phi)}\phi(z)$ for $\phi(z)=\bar g_v^{\,t}(z,x_{\obs(u)})$, or reports that $I\cap\operatorname{dom}(\phi)$ is empty.

\begin{proposition}\label{prop:computation}
Assume the real-arithmetic model and unit-cost access to leaf rewards, feasibility tests, and the interval-supremum oracle.
Then, the policy of Algorithm~\ref{alg:main} in round $t$ can be executed in time polynomial in $n$, $t$, $|V|$, and $|A|$.
\end{proposition}
\begin{proof}
Since $N_{i,t}\le t-1$ for every $i$, preprocessing each sample sequence by sorting and finding distinct values and multiplicities takes time polynomial in $n$ and $t$.
Upon reaching node $u$, the algorithm computes $\bar h_{(u,v)}^{\,t}$ for each feasible action $v$ at the current observations.
For an edge with no observation, this is a leaf-reward evaluation.
For a revelation edge with fewer than 16 samples, this is one oracle query with $I=[0,1]$.
Otherwise, the algorithm obtains $\bar\phi_j$ for the at most $2N_{i,t}+1$ cells from the oracle and computes the optimal value of the LP of Proposition~\ref{prop:local-lp}.
In terms of the cumulative masses $q_{\le j}$, the objective and every constraint of this LP involve only differences $q_{\le c}-q_{\le a-1}$ with $1\le a\le c\le M$.
Hence, the LP maximizes a linear function of the node potentials $q_{\le0},\ldots,q_{\le M}$ subject to upper bounds on their differences, which is the dual of an uncapacitated minimum-cost flow problem with $M+1$ nodes and $O(M^2)$ arcs.
By strong duality, the optimal value of the LP equals the minimum cost of this flow problem.
The potentials given by the empirical distribution are feasible, and replacing the arc costs by the reduced costs with respect to them makes the costs nonnegative without changing the optimal flows.
Then, the strongly polynomial algorithm of Orlin~\cite{Orlin93} computes the minimum cost with $O(M^3\log M)$ arithmetic operations and comparisons.
Since $M\le2N_{i,t}+1\le2t-1$, the computation for each edge, including the $O(t^2)$ arc costs, takes $O(t^3\log t)$ time.
The algorithm then chooses a maximizing feasible action by comparing these values.

Since the nodes visited in a round lie on a single root-to-leaf path, each of the at most $|V|$ nodes is visited at most once, and at most $|A|$ outgoing edges are examined in total, which proves the claim.
\end{proof}

\subsection{Implementing the Interval-Supremum Oracle}\label{subsec:states}
The optimistic DP cannot simply be computed by backward induction, since each optimistic continuation value $\bar g_v^{\,t}$ is a function of the observations $x_{\obs(v)}$, which can take exponentially or infinitely many values.
The interval-supremum oracle hides this difficulty, and each query takes the supremum of the function $\phi(z)=\bar g_v^{\,t}(z,x_{\obs(u)})$ of one variable over $I\cap\operatorname{dom}(\phi)$.
We first reduce a query to finitely many evaluations of $\phi$ and then show how to compute these evaluations.

A query asks for the supremum of $\phi$ over $I\cap\operatorname{dom}(\phi)$, which can be an infinite set, and in the following cases, finitely many evaluations of $\phi$ suffice.
If the domain of coordinate $i$ is finite, a query reduces to evaluating $\phi$ at the points of the domain in $I$.
If the domain is $[0,1]$, $\phi$ is continuous, and $[0,1]$ can be divided into finitely many known intervals on each of which $\phi$ is monotone, a query reduces to evaluating $\phi$ at the endpoints of the closure of $I$ and at the division points in $I$.
If the domain is $[0,1]$ and $\phi$ is continuous and quasi-convex, for example, monotone, the two endpoints of the closure of $I$ suffice, even if the minimizer of $\phi$ is unknown.

If the domain is $[0,1]$ and the optimistic continuation values are $L$-Lipschitz in the revealed value, a query can be approximated within $\eta$ by the maximum of $\phi$ over a grid of the closure of $I$ that contains its endpoints and has spacing at most $\eta/L$.
Each of the at most $n$ revelation edges below a node then adds an error of at most $\eta$, and changing every coefficient of the LP of Proposition~\ref{prop:local-lp} by at most the same amount changes its optimal value by at most this amount since $\sum_jq_j=1$.
Hence, the computed values are within $n\eta$ of the optimistic continuation values, and choosing children by them loses at most $2n\eta$ in the optimistic continuation value at each of the at most $n+1$ stages.
For $\eta=1/(4n(n+1)T)$, these losses add at most $1/2$ to the cumulative gap $\Delta$ in the proof of Theorem~\ref{thm:local}, and the regret bounds increase by at most 1, also in the refinements in which $\Delta$ enters the variance budget.

It remains to compute the evaluations.
Each value of $\phi$ is an optimistic continuation value at $v$ after an additional observation, and computing it requires the optimistic continuation values below $v$ for all further observations, whose number can grow exponentially in $n$.
As in a DP for known distributions, this growth disappears when the past observations matter only through a small state.
For example, in stochastic knapsack (in Section~\ref{subsec:scheduling}) with integer sizes and an integer capacity $B$, the past affects the future only through the set of inserted items and the remaining capacity, and the rewards of the inserted items are added to every subsequent leaf.
With nodes representing sets of inserted items, as in the directed acyclic graph of the footnote in Section~\ref{sec:problem}, the optimistic continuation value at a node is the reward collected so far plus a function of the remaining capacity, and each evaluation of $\phi$ needs this function only at the at most $B+1$ capacities.

Formally, a \emph{state representation} consists of sets $\Pi_u$ for $u\in V$ and maps $\pi_u$ from the observations $x_{\obs(u)}$ to $\Pi_u$ with the following properties.
\begin{enumerate}[label=(\alph*),nosep]
\item The state is updated by the revealed value, that is, each revelation edge $e=(u,v)$ revealing a coordinate $i$ has a map $\pi_e$ with $\pi_v(z,x_{\obs(u)})=\pi_e(\pi_u(x_{\obs(u)}),z)$, and each stopping edge $(u,v)$ satisfies $\pi_v=\pi_u$.
\item The feasible actions depend only on the state, that is, $\mathcal A_u(x_{\obs(u)})=\widetilde{\mathcal A}_u(\pi_u(x_{\obs(u)}))$ for some sets $\widetilde{\mathcal A}_u(\pi)\subseteq\child(u)$.
\item Each leaf reward is a reward of the final state plus rewards collected on the way, that is, $f_\ell(x_{\obs(\ell)})$ is $\beta_\ell(\pi_\ell(x_{\obs(\ell)}))$ plus the sum of $\gamma_e(\pi_u(x_{\obs(u)}),x_i)$ over the revelation edges $e=(u,v)$ on the path to $\ell$, where $i$ is the coordinate revealed by $e$.
\end{enumerate}
For knapsack, the state is the remaining capacity, or a symbol for an overflow, $\gamma_e$ is the normalized reward of the inserted item if it fits, and $\beta_\ell=0$.

The optimistic DP then runs on pairs of a node and a state.
Let $\Phi_\ell=\beta_\ell$ at leaves, and at a non-leaf node $u$, let $\Phi_u(\pi)$ be the maximum of $\Psi_{(u,v)}(\pi)$ over $v\in\widetilde{\mathcal A}_u(\pi)$, where $\Psi_{(u,v)}(\pi)=\Phi_v(\pi)$ for a stopping edge and
\[
\Psi_e(\pi)=\sup_{Q\in\CC_{i,t}}\E_{X\sim Q}[\psi_{e,\pi}(X)],
\qquad
\psi_{e,\pi}(z)=\gamma_e(\pi,z)+\Phi_v(\pi_e(\pi,z)),
\]
for a revelation edge $e=(u,v)$ revealing $i$.
Let $\kappa_u(x_{\obs(u)})$ be the sum of the rewards $\gamma_e$ collected on the path to $u$.
Since expectations, suprema, and maxima commute with adding $\kappa_u$, backward induction shows that $\bar g_u^{\,t}(x_{\obs(u)})=\kappa_u(x_{\obs(u)})+\Phi_u(\pi_u(x_{\obs(u)}))$.

By Proposition~\ref{prop:local-lp}, each $\Psi_e(\pi)$ is determined by the suprema of $\psi_{e,\pi}$ over the cells and over the domain.
We call a finite set $\mathcal Z_e(\pi)$ of points of the domain of coordinate $i$ \emph{sufficient} if each of these suprema is the maximum of $\psi_{e,\pi}$ over the points of $\mathcal Z_e(\pi)$ in the closure of the corresponding set.
By the reductions above, the domain is sufficient if it is finite, and $\{0,1\}$ together with the sample points of coordinate $i$ is sufficient if the domain is $[0,1]$ and $\psi_{e,\pi}$ is continuous and quasi-convex.
Computing $\Psi_e(\pi)$ then requires $\Phi_v$ only at the states $\pi_e(\pi,z)$ with $z\in\mathcal Z_e(\pi)$.
Repeating this argument, the recursion evaluates $\Phi$ only at the pairs \emph{reachable} from $(u,\pi)$, which are obtained by moving along feasible actions, keeping the state on stopping edges and applying $\pi_e(\cdot,z)$ with $z$ in the sufficient sets on revelation edges.
Their states are of the form $\pi_v(x_{\obs(v)})$, since the sufficient sets consist of points of the domains.
Let $K_\Pi$ be the largest number of pairs reachable from a pair
$(u,\pi_u(x_{\obs(u)}))$ over all nodes $u$ and observations
$x_{\obs(u)}$.

For the following proposition, we work in the real-arithmetic model
and assume that evaluating $\pi_r$, $\pi_e$, $\gamma_e$, and $\beta_\ell$,
testing whether $v\in\widetilde{\mathcal A}_u(\pi)$,
listing the points of $\mathcal Z_e(\pi)$,
and testing whether two states are equal all take unit time.

\begin{proposition}\label{prop:states}
Suppose that the problem has a state representation with known
sufficient sets $\mathcal Z_e(\pi)$ for every state
$\pi=\pi_u(x_{\obs(u)})$.
Then, the policy of Algorithm~\ref{alg:main} in round $t$
can be executed in time polynomial in $n$, $t$, $|A|$, $K_\Pi$,
and the sizes of the sets $\mathcal Z_e(\pi)$.
\end{proposition}
\begin{proof}
At each of the at most $n+1$ visited non-leaf nodes $u$ with state $\pi$, the policy compares the values $\Psi_{(u,v)}(\pi)$, which by the recursion above need $\Phi$ only at the pairs reachable from $(u,\pi)$.
We compute $\Phi$ at each of these at most $K_\Pi$ pairs once, by backward induction along the edges, and we detect repeated pairs by testing states for equality.
For a revelation edge $e$ and a state $\pi=\pi_u(x_{\obs(u)})$, the value $\Psi_e(\pi)$ is the optimal value of the LP of Proposition~\ref{prop:local-lp} for $\phi$ minus $\kappa_u(x_{\obs(u)})$, which is the optimal value of the same LP with coefficients for $\psi_{e,\pi}$.
These coefficients are the maxima of $\psi_{e,\pi}$ over the points of $\mathcal Z_e(\pi)$ in the closures of the cells, or one maximum over the domain if $N_{i,t}<16$, and placing these points into the cells takes time polynomial in $t$ and $|\mathcal Z_e(\pi)|$.
Since each LP takes $O(t^3\log t)$ time as in the proof of Proposition~\ref{prop:computation}, each reachable pair takes time polynomial in $t$, $|A|$, and the sizes of the sets $\mathcal Z_e(\pi)$.
\end{proof}

We now apply Proposition~\ref{prop:states} to examples.
If every domain is finite, the domains are sufficient, and Proposition~\ref{prop:states} shows that each round runs in time polynomial in $n$, $t$, $|A|$, $K_\Pi$, and the sizes of the domains.
If every domain is $[0,1]$ and the optimistic continuation values are $L$-Lipschitz in the revealed value, the grids of spacing $\eta/L$ in the closures of the cells with $\eta=1/(4n(n+1)T)$ replace the sufficient sets with an error of at most $\eta$ in each supremum.
As shown above, the regret bounds then increase by at most 1, and the proof of Proposition~\ref{prop:states} shows that each round runs in time polynomial in $n$, $T$, $|A|$, $K_\Pi$, and $L$, where $K_\Pi$ counts the pairs reachable through the grid points.

For overbooking with group reservations in Section~\ref{subsec:overbooking}, let the state at a node be the total number of arrivals in the groups revealed so far, which is 0 at the root and at the leaf for $U=\emptyset$.
With $\gamma_e(\pi,z)=(a_i-\rho_ib_i+\rho_ib_iz)/\Gamma$ for the edge revealing group $i$, $\pi_e(\pi,z)=\pi+b_iz$, and $\beta_\ell(\pi)=(C-c(\pi-B)_+)/\Gamma$, the domains are sufficient, and every reachable state is an integer from 0 to $\bar b$.
Since $|V|\le n|\mathcal U|+2$, we have $K_\Pi\le(n|\mathcal U|+2)(\bar b+1)$, and Proposition~\ref{prop:states} shows that each round runs in time polynomial in $n$, $t$, $|\mathcal U|$, and $\bar b$.

For optimal stopping with loss aversion in Section~\ref{subsec:behavioral}, let $\omega(\xi,z)=(z-\lambda(\xi-z)_++\lambda)/(1+\lambda)$ be the normalized reward for stopping at value $z$ when $\xi$ is the maximum of the past values.
The state at $u_i$ and $\ell_i$ is the pair $(\xi,x_i)$ with $\xi=\max_{j<i}x_j$, the transition is $\pi_e((\xi,x),z)=(\max\{\xi,x\},z)$, $\gamma_e=0$, and $\beta_{\ell_i}=\omega$.
Then, $\Phi_{u_n}=\omega$ and, for $i<n$, $\Phi_{u_i}(\xi,z)=\max\{\omega(\xi,z),H_{i+1}(\max\{\xi,z\})\}$ with $H_{i+1}(\xi)=\sup_{Q\in\CC_{i+1,t}}\E_{X\sim Q}[\Phi_{u_{i+1}}(\xi,X)]$.
Since $\omega(\cdot,z)$ is nonincreasing and $\lambda/(1+\lambda)$-Lipschitz for every $z$, and composition with $\xi\mapsto\max\{\xi,z\}$, maxima, expectations, and suprema preserve these properties with the same constant, every $H_i$ is nonincreasing and continuous, and every $\Phi_{u_i}(\xi,\cdot)$ is continuous and quasi-convex.
Hence, if each domain is $[0,1]$ or finite, the set $\mathcal E$ consisting of 0, 1, the sample values of all coordinates, and the points of the finite domains has the property that its points in the domain of the revealed coordinate form a sufficient set for every edge and state.
For a state $(\xi,z_0)$, let $w_0=\max\{\xi,z_0\}$.
Since $\max\{w,z\}\in\mathcal E\cup\{w_0\}$ for $w\in\mathcal E\cup\{w_0\}$ and $z\in\mathcal E$, every state reachable from $(\xi,z_0)$ other than $(\xi,z_0)$ itself lies in $(\mathcal E\cup\{w_0\})\times\mathcal E$.
Hence, $K_\Pi\le2n(|\mathcal E|+1)^2$, since the tree has $2n-1$ nodes, and each round runs in polynomial time, since $|\mathcal E|$ is polynomial in $n$, $t$, and the sizes of the finite domains.

\section{Efficient Learning for SBFE with Threshold Functions}
\label{app:sbfe-threshold}

In this section, we give a polynomial-time learning algorithm for
SBFE with threshold functions.
Consider
\[
F(x)=\ind\{x_1+\cdots+x_n\ge\theta\},
\qquad \theta\in[n],
\]
with positive testing costs $c_1,\ldots,c_n$, and let
$\Gamma=\sum_{i=1}^n c_i$.

At the beginning of each round $t$, form the confidence sets
$\CC_{i,t}$ as in Section~\ref{subsec:optimistic-dp}, and let
\[
\ell_i=\min_{Q\in\CC_{i,t}}Q(\{1\}),
\qquad
u_i=\max_{Q\in\CC_{i,t}}Q(\{1\}).
\]
Order the coordinates by increasing $c_i/u_i$ and, separately,
by increasing $c_i/(1-\ell_i)$.
We use the convention $c_i/0=\infty$ and break ties by index
in both orders.
These orders remain fixed throughout the round.

While $F$ is not yet determined, let $m$ be the number of untested
coordinates and let $\theta'$ be the number of additional 1's
needed to establish $F(x)=1$.
Among the untested coordinates, consider the first $\theta'$ in the
$c_i/u_i$ order and the first $m-\theta'+1$ in the
$c_i/(1-\ell_i)$ order.
These two sets intersect because their sizes sum to $m+1>m$.
The algorithm tests any coordinate in their intersection and repeats
until $F$ is determined.
At the end of the round, it adds the observed values to the samples.

\begin{proposition}\label{prop:sbfe-threshold}
For $\delta=1/(T+1)$, the algorithm described above can be implemented
in $O(n^2)$ time per round and has expected regret
$\otil(\Gamma\sqrt{nT})$ in the original testing cost.
\end{proposition}

\begin{proof}
The endpoints $\ell_i$ and $u_i$ for all coordinates can be computed
in $O(n)$ time from the sample counts using
\eqref{eq:radius} and~\eqref{eq:confidence}.
Sorting the two orders once per round and scanning them at each
of the at most $n$ steps takes $O(n^2)$ time.

We next bound the regret.
Write the normalized reward from Section~\ref{subsec:sbfe} as
$f=f_0+f_1$, where
$f_b(\sigma,x)=f(\sigma,x)\ind\{F(x)=b\}$ for $b\in\{0,1\}$.
For $q\in[0,1]^n$, write
$\operatorname{Bernoulli}(q)=\prod_i\operatorname{Bernoulli}(q_i)$,
and define
\[
V_b(q)=\max_\sigma
\E_{X\sim\operatorname{Bernoulli}(q)}[f_b(\sigma,X)].
\]
Then
$\opt(\operatorname{Bernoulli}(q))\le V_0(q)+V_1(q)$.
The increasing orders of $c_i/q_i$ and $c_i/(1-q_i)$ attain
$V_1(q)$ and $V_0(q)$, respectively~\cite{BenDov1981,SalloumBreuer1984}.
This also holds when some probabilities are 0 or 1 by continuity,
since there are finitely many policies.

Fix a round $t$, and write
$\ell=(\ell_1,\ldots,\ell_n)$ and $u=(u_1,\ldots,u_n)$.
Consider a stage with $m$ untested coordinates and remaining
threshold $\theta'$.
On every completion with $F(x)=1$, a fixed order of the remaining
coordinates tests all of its first $\theta'$ coordinates.
Moving any one of these coordinates to the front therefore does not
change the testing cost on such completions.
The same holds for completions with $F(x)=0$ and the first
$m-\theta'+1$ coordinates.
By induction on $m$, the policy $\sigma^t$ of the algorithm has the
same testing cost as the $c_i/u_i$ order on inputs with $F(x)=1$
and as the $c_i/(1-\ell_i)$ order on inputs with $F(x)=0$.
Hence, it attains $V_1(u)$ under $\operatorname{Bernoulli}(u)$
and $V_0(\ell)$ under $\operatorname{Bernoulli}(\ell)$.

For a fixed testing order, $f_1$ is nondecreasing and $f_0$ is
nonincreasing in each $x_i$.
Thus, $V_1$ is nondecreasing and $V_0$ is nonincreasing in each $q_i$.
Let $p_i=\Prb(X_i=1)$ and $p=(p_1,\ldots,p_n)$.
If every $D_i$ belongs to $\CC_{i,t}$, then
$\ell_i\le p_i\le u_i$ for every $i$, so
$V_1(u)\ge V_1(p)$ and $V_0(\ell)\ge V_0(p)$.
Proposition~\ref{prop:single}(b) with $\alpha=1$ and $K=2n$
therefore applies to $f_1$ with
$Q^t=\operatorname{Bernoulli}(u)$ and to $f_0$ with
$Q^t=\operatorname{Bernoulli}(\ell)$.
Adding the two bounds yields
\[
\sum_{t=1}^T
\bigl(V_0(p)+V_1(p)-\E[f(\sigma^t,X^t)]\bigr)
=\otil(\sqrt{nT}).
\]
This bounds the normalized regret since
$\opt(\operatorname{Bernoulli}(p))\le V_0(p)+V_1(p)$.
Multiplying by $\Gamma$ gives the regret bound in the original
testing cost.
\end{proof}

\section{Omitted Proofs}\label{app:proofs}

\begin{proof}[Proof of Proposition~\ref{prop:bias-lower}]
Let $n=2$, let $X_1=1/2$ with probability 1, and let $X_2\in\{0,1\}$ with $\Prb(X_2=0)=q$.
Stopping at the first candidate yields utility $1/2$, and continuing yields expected utility $(1-q)-q\lambda/2=1-(\lambda+2)q/2$.
Let $q_0=1/(\lambda+2)$, $\varepsilon=\frac18\min\{q_0,\sqrt{q_0/T}\}$, and $q_\pm=q_0\pm\varepsilon$.
Continuing is optimal for $q=q_-$, and stopping is optimal for $q=q_+$.
In both instances, the other action loses $(\lambda+2)\varepsilon/2=\frac1{16}\min\{1,\sqrt{(\lambda+2)/T}\}$ in expected utility.

Under full feedback, $X_2$ is revealed at the end of every round regardless of the action, which provides more information than semi-bandit feedback.
Fix a round $t$, and let $P_+$ and $P_-$ be the distributions of the $t-1$ previous samples of $X_2$ for $q=q_+$ and $q=q_-$, respectively.
The action in round $t$ depends only on these samples and the internal randomness of the learning algorithm.
The KL divergence between the distributions of one sample is at most their chi-squared divergence $(q_+-q_-)^2/(q_-(1-q_-))$.
Since $q_-\ge7q_0/8$ and $1-q_-\ge1/2$, we have
\[
\KL(P_+\|P_-)\le\frac{(t-1)(2\varepsilon)^2}{q_-(1-q_-)}\le\frac{64T\varepsilon^2}{7q_0}\le\frac17.
\]
By Pinsker's inequality, $\TV(P_+,P_-)\le1/\sqrt{14}<1/3$.
As in the proof of Lemma~\ref{lem:testing-lower}, the probability of the wrong action in round $t$, averaged over the two instances, is at least $(1-\TV(P_+,P_-))/2\ge1/3$.
Summing over the rounds, the average of the regrets on the two instances is at least $\frac T3\cdot\frac1{16}\min\{1,\sqrt{(\lambda+2)/T}\}=\frac1{48}\min\{T,\sqrt{(\lambda+2)T}\}$, and hence the regret on one of them is at least this value.
\end{proof}

\begin{proof}[Proof of Proposition~\ref{prop:overbooking}]
We verify the conditions of Theorem~\ref{thm:sum} with $s_i=1$, for which $S=n$.
Consider the edge revealing group $i$ in the chain for an accepted set $U$.
Let $\varrho$ be the sum of $a_j-\rho_j(b_j-Y_j)$ over the groups $j$ of $U$ revealed before group $i$, and let $J$ be the set of groups of $U$ revealed after group $i$.
The only leaf below this edge is the leaf for $U$, and with $Y_i=b_ix_i$, its reward is $\kappa+\chi(x_i)+F$, where $\kappa=(\varrho+a_i-\rho_ib_i)/\Gamma$, $\chi(z)=\rho_ib_iz/\Gamma$, and
\[
F=\frac1\Gamma\Bigl(\sum_{j\in J}\bigl(a_j-\rho_j(b_j-Y_j)\bigr)-c\Bigl(\sum_{j\in U}Y_j-B\Bigr)_++C\Bigr).
\]
The function $\chi$ has level-set complexity at most 1 because it is nondecreasing.
Moreover, $\chi,F\in[0,1]$ by $\rho_jb_j\le a_j$ and $\sum_{j\in U}Y_j\le\bar b$.
The residual reward $F$ is nonincreasing in $x_i$, and the sets of feasible actions do not depend on $x_i$.
Finally, the immediate rewards sum to $\sum_{j\in U}\rho_jY_j/\Gamma\le1$.
Theorem~\ref{thm:sum} and rescaling by $\Gamma$ imply the claim.
\end{proof}

\begin{proof}[Proof of Lemma~\ref{lem:interval-bernstein}]
We apply Maurer and Pontil's theorem~\cite[Theorem~6]{mp2009} to the function class $\{\ind_I:I\in\II_1\}\cup\{1-\ind_I:I\in\II_1\}$.
Its uniform covering number at scale $1/m$ on $2m$ points is bounded by the number of distinct value patterns.
On ordered points $x_1\le\cdots\le x_{2m}$, an interval selects a consecutive block, yielding at most $(2m+1)^2$ patterns.
Including the complementary indicators yields at most $2(2m+1)^2$ patterns, which combines with the factor 10 in their theorem to produce the numerator in $\beta$.
The sample variance of either indicator is $m\widehat P_m(I)(1-\widehat P_m(I))/(m-1)$, yielding the variance term in~\eqref{eq:mp}.
\end{proof}

\bibliographystyle{alphaurl}
\bibliography{ref}

@article{Alaei14,
  author = {Alaei, Saeed},
  title = {{Bayesian} Combinatorial Auctions: Expanding Single Buyer Mechanisms to Many Buyers},
  journal = {SIAM Journal on Computing},
  volume = {43},
  number = {2},
  pages = {930--972},
  year = {2014},
}

@article{asadpour2016maximizing,
  author = {Asadpour, Arash and Nazerzadeh, Hamid},
  title = {Maximizing Stochastic Monotone Submodular Functions},
  journal = {Management Science},
  volume = {62},
  number = {8},
  pages = {2374--2391},
  year = {2016},
}

@inproceedings{AzarOM17,
  author = {Azar, Mohammad Gheshlaghi and Osband, Ian and Munos, R{\'e}mi},
  title = {Minimax Regret Bounds for Reinforcement Learning},
  booktitle = {Proceedings of the 34th International Conference on Machine Learning (ICML)},
  pages = {263--272},
  year = {2017}
}

@article{BenDov1981,
  author = {Ben-Dov, Yosi},
  title = {Optimal Testing Procedures for Special Structures of Coherent Systems},
  journal = {Management Science},
  volume = {27},
  number = {12},
  pages = {1410--1420},
  year = {1981},
}

@book{cesa2006prediction,
  author = {Cesa-Bianchi, Nicol{\`o} and Lugosi, G{\'a}bor},
  title = {Prediction, Learning, and Games},
  publisher = {Cambridge University Press},
  year = {2006},
}

@inproceedings{icml/ChenWY13,
  author = {Chen, Wei and Wang, Yajun and Yuan, Yang},
  title = {Combinatorial Multi-Armed Bandit: General Framework and Applications},
  booktitle = {Proceedings of the 30th International Conference on Machine Learning (ICML)},
  pages = {151--159},
  year = {2013}
}

@inproceedings{chen2016general,
  author = {Chen, Wei and Hu, Wei and Li, Fu and Li, Jian and Liu, Yu and Lu, Pinyan},
  title = {Combinatorial Multi-Armed Bandit with General Reward Functions},
  booktitle = {Proceedings of the 30th Conference on Neural Information Processing System (NIPS)},
  pages = {1659--1667},
  year = {2016}
}

@article{chen2016cmab,
  author = {Chen, Wei and Wang, Yajun and Yuan, Yang and Wang, Qinshi},
  title = {Combinatorial Multi-Armed Bandit and Its Extension to Probabilistically Triggered Arms},
  journal = {Journal of Machine Learning Research},
  volume = {17},
  number = {50},
  pages = {1--33},
  year = {2016}
}

@inproceedings{correa2019prophet,
  author = {Correa, Jos{\'e} and D{\"u}tting, Paul and Fischer, Felix and Schewior, Kevin},
  title = {Prophet Inequalities for {i.i.d.} Random Variables from an Unknown Distribution},
  booktitle = {Proceedings of the 20th ACM Conference on Economics and Computation (EC)},
  pages = {3--17},
  year = {2019},
}

@phdthesis{dean2005thesis,
  author = {Dean, Brian C.},
  title = {Approximation Algorithms for Stochastic Scheduling Problems},
  school = {Massachusetts Institute of Technology},
  year = {2005}
}

@article{DeanGV08,
  author = {Dean, Brian C. and Goemans, Michel X. and Vondr{\'a}k, Jan},
  title = {Approximating the Stochastic Knapsack Problem: The Benefit of Adaptivity},
  journal = {Mathematics of Operations Research},
  volume = {33},
  number = {4},
  pages = {945--964},
  year = {2008},
}

@inproceedings{dhk2014,
  author = {Deshpande, Amol and Hellerstein, Lisa and Kletenik, Devorah},
  title = {Approximation Algorithms for Stochastic {Boolean} Function Evaluation and Stochastic Submodular Set Cover},
  booktitle = {Proceedings of the 25th ACM-SIAM Symposium on Discrete Algorithms (SODA)},
  pages = {1453--1466},
  year = {2014},
}

@inproceedings{esfandiari2020prophets,
  author = {Esfandiari, Hossein and Hajiaghayi, MohammadTaghi and Lucier, Brendan and Mitzenmacher, Michael},
  title = {Prophets, Secretaries, and Maximizing the Probability of Choosing the Best},
  booktitle = {Proceedings of the 23rd International Conference on Artificial Intelligence and Statistics (AISTATS)},
  pages = {3717--3727},
  year = {2020}
}

@inproceedings{Gatmiry2024,
  author = {Gatmiry, Khashayar and Kesselheim, Thomas and Singla, Sahil and Wang, Yifan},
  title = {Bandit Algorithms for Prophet Inequality and {Pandora}'s Box},
  booktitle = {Proceedings of the 35th ACM-SIAM Symposium on Discrete Algorithms (SODA)},
  pages = {462--500},
  year = {2024},
}

@article{ghuge2026semibandit,
  author = {Ghuge, Rohan and Zhuo, Zhengjia and Agarwal, Arpit and Nagarajan, Viswanath},
  title = {Semibandit Learning for Monotone Stochastic Optimization},
  journal = {Operations Research},
  year = {2026},
}

@article{gilbert1966recognizing,
  author = {Gilbert, John P. and Mosteller, Frederick},
  title = {Recognizing the Maximum of a Sequence},
  journal = {Journal of the American Statistical Association},
  volume = {61},
  number = {313},
  pages = {35--73},
  year = {1966},
}

@inproceedings{guo2019settling,
  author = {Guo, Chenghao and Huang, Zhiyi and Zhang, Xinzhi},
  title = {Settling the Sample Complexity of Single-Parameter Revenue Maximization},
  booktitle = {Proceedings of the 51st ACM Symposium on Theory of Computing (STOC)},
  pages = {662--673},
  year = {2019},
}

@inproceedings{pmlr-v134-guo21a,
  author = {Guo, Chenghao and Huang, Zhiyi and Tang, Zhihao Gavin and Zhang, Xinzhi},
  title = {Generalizing Complex Hypotheses on Product Distributions: Auctions, Prophet Inequalities, and {Pandora}'s Problem},
  booktitle = {Proceedings of the 34th Conference on Learning Theory (COLT)},
  pages = {2248--2288},
  year = {2021}
}

@inproceedings{GN13,
  author = {Gupta, Anupam and Nagarajan, Viswanath},
  title = {A Stochastic Probing Problem with Applications},
  booktitle = {Proceedings of the 16th International Conference on Integer Programming and Combinatorial Optimization (IPCO)},
  pages = {205--216},
  year = {2013},
}

@inproceedings{GuptaKMR11,
  author = {Gupta, Anupam and Krishnaswamy, Ravishankar and Molinaro, Marco and Ravi, R.},
  title = {Approximation Algorithms for Correlated Knapsacks and Non-Martingale Bandits},
  booktitle = {Proceedings of the 52nd IEEE Symposium on Foundations of Computer Science (FOCS)},
  pages = {827--836},
  year = {2011},
}

@misc{harada2025repeated,
  author = {Harada, Tsubasa and Kawase, Yasushi and Sumita, Hanna},
  title = {Online Algorithms for Repeated Optimal Stopping: Balancing Baseline Guarantees and Regret},
  howpublished = {arXiv:2511.04484v2},
  year = {2025}
}

@misc{hg2026,
  author = {Hu, Tao and Ghuge, Rohan},
  title = {Near-Optimal Regret Bounds for Monotone Stochastic Optimization under Semi-Bandit Feedback},
  howpublished = {arXiv:2610.17320v1},
  year = {2026}
}

@article{jaksch2010near,
  author = {Jaksch, Thomas and Ortner, Ronald and Auer, Peter},
  title = {Near-Optimal Regret Bounds for Reinforcement Learning},
  journal = {Journal of Machine Learning Research},
  volume = {11},
  number = {51},
  pages = {1563--1600},
  year = {2010}
}

@article{kleinberg2022optimal,
  author = {Kleinberg, Jon and Kleinberg, Robert and Oren, Sigal},
  title = {Optimal Stopping with Behaviorally Biased Agents: The Role of Loss Aversion and Changing Reference Points},
  journal = {Games and Economic Behavior},
  volume = {133},
  pages = {282--299},
  year = {2022},
}

@inproceedings{kleinberg2012matroid,
  author = {Kleinberg, Robert and Weinberg, S. Matthew},
  title = {Matroid Prophet Inequalities},
  booktitle = {Proceedings of the 44th ACM Symposium on Theory of Computing (STOC)},
  pages = {123--136},
  year = {2012},
}

@article{Krengel1977-uu,
  author = {Krengel, Ulrich and Sucheston, Louis},
  title = {Semiamarts and Finite Values},
  journal = {Bulletin of the American Mathematical Society},
  volume = {83},
  number = {4},
  pages = {745--747},
  year = {1977},
}

@inproceedings{KvetonWAS15,
  author = {Kveton, Branislav and Wen, Zheng and Ashkan, Azin and Szepesv{\'a}ri, Csaba},
  title = {Tight Regret Bounds for Stochastic Combinatorial Semi-Bandits},
  booktitle = {Proceedings of the 18th International Conference on Artificial Intelligence and Statistics (AISTATS)},
  pages = {535--543},
  year = {2015}
}

@book{lattimore2020bandit,
  author = {Lattimore, Tor and Szepesv{\'a}ri, Csaba},
  title = {Bandit Algorithms},
  publisher = {Cambridge University Press},
  year = {2020},
}

@inproceedings{liu2025improved,
  author = {Liu, Junyan and Chen, Ziyun and Wang, Kun and Luo, Haipeng and Ratliff, Lillian J.},
  title = {Improved Regret and Contextual Linear Extension for {Pandora}'s Box and Prophet Inequality},
  booktitle = {Proceedings of the 39th International Conference on Neural Information Processing Systems (NeurIPS)},
  pages = {46877--46914},
  year = {2025},
}

@article{Ma18,
  author = {Ma, Will},
  title = {Improvements and Generalizations of Stochastic Knapsack and {Markovian} Bandits Approximation Algorithms},
  journal = {Mathematics of Operations Research},
  volume = {43},
  number = {3},
  pages = {789--812},
  year = {2018},
}

@inproceedings{mp2009,
  author = {Maurer, Andreas and Pontil, Massimiliano},
  title = {Empirical {Bernstein} Bounds and Sample Variance Penalization},
  booktitle = {Proceedings of the 22nd Conference on Learning Theory (COLT)},
  year = {2009}
}

@article{Orlin93,
  author = {Orlin, James B.},
  title = {A Faster Strongly Polynomial Minimum Cost Flow Algorithm},
  journal = {Operations Research},
  volume = {41},
  number = {2},
  pages = {338--350},
  year = {1993},
}

@article{SalloumBreuer1984,
  author = {Salloum, Salam and Breuer, Melvin A.},
  title = {An Optimum Testing Algorithm for Some Symmetric Coherent Systems},
  journal = {Journal of Mathematical Analysis and Applications},
  volume = {101},
  number = {1},
  pages = {170--194},
  year = {1984},
}

@article{samuel1984comparison,
  author = {Samuel-Cahn, Ester},
  title = {Comparison of Threshold Stop Rules and Maximum for Independent Nonnegative Random Variables},
  journal = {The Annals of Probability},
  volume = {12},
  number = {4},
  pages = {1213--1216},
  year = {1984},
}

@article{subramanian1999,
  author = {Subramanian, Janakiram and Stidham, Jr., Shaler and Lautenbacher, Conrad J.},
  title = {Airline Yield Management with Overbooking, Cancellations, and No-Shows},
  journal = {Transportation Science},
  volume = {33},
  number = {2},
  pages = {147--167},
  year = {1999},
}

@inproceedings{tian2020factored,
  author = {Tian, Yi and Qian, Jian and Sra, Suvrit},
  title = {Towards Minimax Optimal Reinforcement Learning in Factored {Markov} Decision Processes},
  booktitle = {Proceedings of the 34th Conference on Neural Information Processing System (NeurIPS)},
  pages = {19896--19907},
  year = {2020}
}

@article{weitzman1979,
  author = {Weitzman, Martin L.},
  title = {Optimal Search for the Best Alternative},
  journal = {Econometrica},
  volume = {47},
  number = {3},
  pages = {641--654},
  year = {1979},
}

@inproceedings{YoshinagaK25,
  author = {Yoshinaga, Toru and Kawase, Yasushi},
  title = {Online Contention Resolution Schemes for Size-Stochastic Knapsacks},
  booktitle = {Proceedings of the 19th International Conference and Workshops on Algorithms and Computation (WALCOM)},
  pages = {393--408},
  year = {2025},
}

@misc{zhuo2026nearoptimal,
  author = {Zhuo, Zhengjia and Nagarajan, Viswanath},
  title = {Near-Optimal Regret for Semi-Bandit Learning of Monotone Stochastic Problems},
  howpublished = {arXiv:2610.24446v1},
  year = {2026}
}
\end{document}